\documentclass[runningheads]{llncs}

\usepackage{eccv}

\usepackage{eccvabbrv}

\usepackage{graphicx}
\usepackage{booktabs}

\usepackage{pifont}

\newcommand{\rom}[1]{\uppercase\expandafter{\romannumeral #1\relax}}

\usepackage{xcolor}

\usepackage{calc}

\usepackage[accsupp]{axessibility}  
\usepackage{colortbl}
\usepackage{tabulary}
\usepackage{tabularx}
\usepackage{etoolbox}
\usepackage{multirow}
\usepackage{cuted}
\usepackage[percentage]{overpic}
\usepackage{booktabs}
\usepackage{pgfplots}
\pgfplotsset{compat=newest}
\usepackage{algorithm}
\usepackage{algpseudocode}
\usepackage{amsmath}
\usepackage{bbding}

\usepackage{stackrel}

\usepackage{tikz}
\usetikzlibrary{calc}
\usetikzlibrary{decorations.pathmorphing}
\usetikzlibrary{patterns,arrows,automata,positioning,decorations,shapes,calc}

\tikzstyle{normalvertex}=[circle,fill=white,draw=black]
\tikzstyle{emptyvertex}=[draw,circle,minimum size=7pt,inner sep=0pt]
\tikzstyle{tinyvertex}=[draw,circle,minimum size=3pt,inner sep=0pt]
\tikzstyle{thickedge}=[draw,gray!60,line width=1.6pt,-]

\usepackage{pgfplots}
\usepgfplotslibrary{colorbrewer}

\pgfdeclarelayer{background}
\pgfsetlayers{background,main}

\usetikzlibrary{shapes}
\usetikzlibrary{arrows.meta,backgrounds,automata, chains}
\usetikzlibrary{positioning,shapes,matrix,calc}
\tikzstyle{vertex}=[circle, draw, fill=gray!80!white,thick,scale=1.2]
\tikzstyle{edge}=[draw=black, thick,-]

\usepackage{bm}

\definecolor{purple}{RGB}{147,7,204}
\definecolor{blue}{RGB}{10,153,201}
\definecolor{orange}{RGB}{254,128,41}
\definecolor{gray}{RGB}{239,240,241}
\definecolor{pink}{RGB}{254,15,127}
\definecolor{green}{RGB}{140,211,89}

\definecolor{color1}{RGB}{254,15,127}
\definecolor{color2}{RGB}{10,153,201}
\definecolor{color3}{RGB}{194,145,162}
\definecolor{color4}{RGB}{254,128,41}
\definecolor{color5}{RGB}{254,191,185}
\definecolor{color6}{RGB}{110,231,169}
\definecolor{color7}{RGB}{245,221,66}

\newcommand{\shrec}{\textsc{Shrec16}}
\newcommand{\tosca}{\textsc{Tosca}}
\newcommand{\cuts}{\textsc{Cuts}}
\newcommand{\holes}{\textsc{Holes}}
\newcommand{\cutsTwentyFour}{\textsc{Cuts24}}

\newcommand{\pfaust}{\textsc{Pfaust}}

\newcommand{\pfaustM}{\textsc{Pfaust-M}}
\newcommand{\pfaustH}{\textsc{Pfaust-H}}

\def\pathOurs{figs/ours/}
\def\srcEnd{_M}
\def\trgtEnd{_N}

\usepackage{hyperref}

\usepackage{orcidlink}

\begin{document}
\setkeys{Gin}{keepaspectratio}
\title{Hyper-Network Neural Functional Maps for Unsupervised Robust 3D Shape Matching} 

\titlerunning{Hyper-Network Neural Functional Maps}

\author{Dongliang Cao\inst{1}\orcidlink{0000-0002-6505-6465} \and
Florian Bernard\inst{1}\orcidlink{0009-0008-1137-0003} 
}

\authorrunning{D.~Cao, F.~Bernard.}

\institute{University of Bonn, Germany}

\maketitle

\begin{abstract}
Functional maps are the cornerstone of recent non-rigid 3D shape matching methods due to their efficiency and performance. However, existing methods struggle with challenging scenarios, such as partiality, topological noise, and raw point clouds. A primary bottleneck is that significant intrinsic distortion prevents truncated spectral bases from being accurately aligned via linear transformations (i.e., functional maps). To address this, we introduce a \textit{hyper-network that predicts non-linear neural functional maps (NFM)}, learned in an unsupervised manner, to better align spectral bases. Specifically, we model the NFM as an MLP with skip-connection to refine standard FM and employ a hyper-network to predict its weights, conditioned on standard FM. Our framework is trained using a novel unsupervised spectral alignment loss. Experiments demonstrate that our approach can be seamlessly integrated into state-of-the-art unsupervised deep functional map pipelines, substantially improving matching accuracy in demanding scenarios.  
\end{abstract}   
\begin{figure}[bht!]
    \centering
    \includegraphics[width=\textwidth]{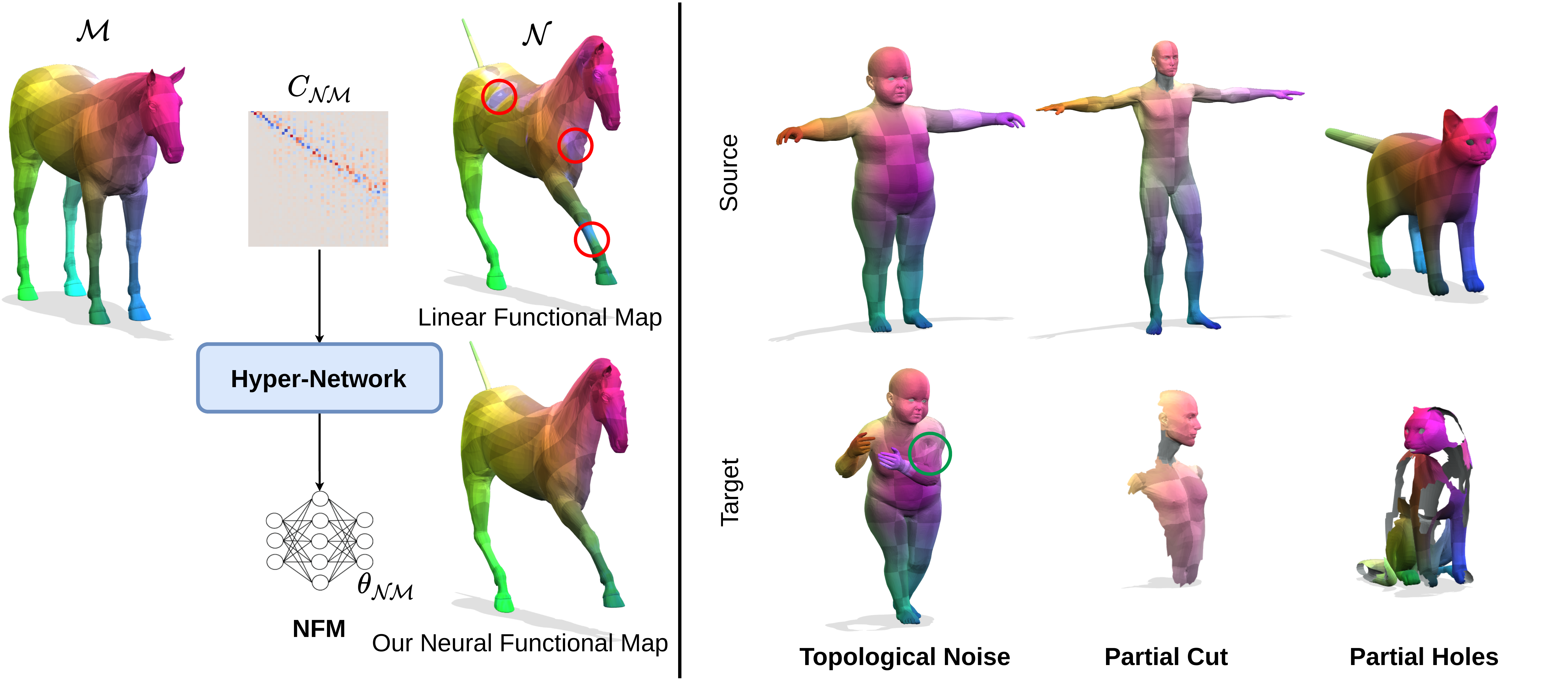}
    \caption{Standard functional maps leverage a \emph{linear} transformation and often fail to establish accurate correspondences when faced with significant intrinsic distortions -- this is because a linear transformation is insufficient to align truncated spectral bases under severe partiality or topological noise. To overcome this, we introduce \textit{non-linear neural functional maps (NFM)}, which are dynamically predicted by a hyper-network {conditioned on standard FM}. Our framework enables a more flexible and robust basis alignment, allowing our method to recover high-quality vertex-wise correspondences even in the presence of topological noise and extreme missing parts.
    }
    \label{fig:teaser}
\end{figure}

\section{Introduction}
\label{sec:intro}

Establishing correspondences between non-rigid 3D shapes is a fundamental problem in computer vision and computer graphics. Its utility spans a diverse range of applications, from texture and pose transfer~\cite{dinh2005texture, song20213d, song2023unsupervised} to statistical shape modeling~\cite{loper2015smpl, li2017flame, egger20203d}. Despite being a long-standing challenge with decades of extensive study~\cite{deng2022survey, sahilliouglu2020recent}, achieving accurate vertex-wise correspondence still remains a challenging task. This is particularly evident when dealing with topological noise and partiality. Such distortions are nearly ubiquitous in real-world 3D scanning data but difficult for traditional algorithms~\cite{ovsjanikov2012functional,ovsjanikov2010one} to tackle.

\begin{figure}[bht!]
    \centering
    \includegraphics[width=\textwidth]{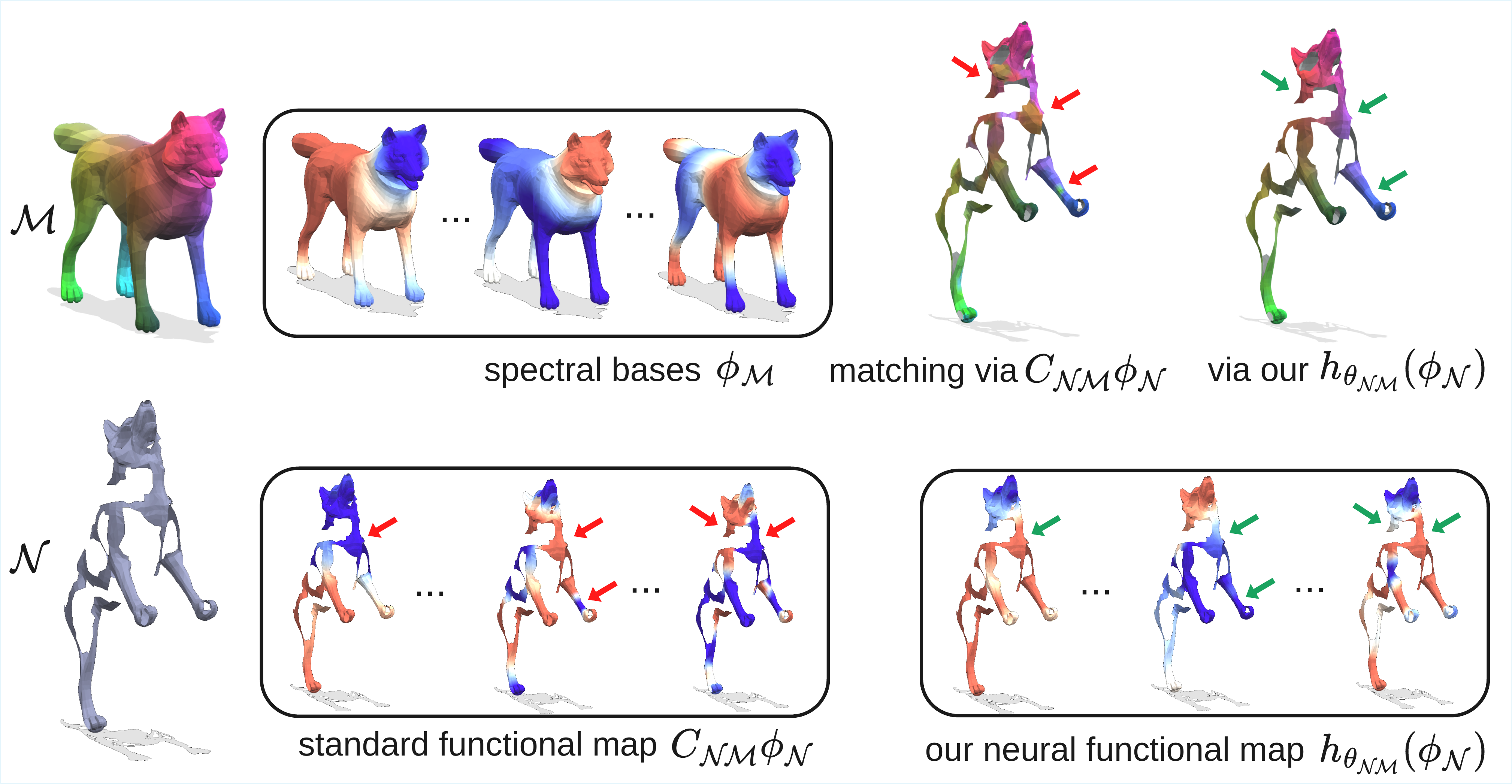}
    \caption{Comparison between standard and neural functional maps on a challenging partial shape matching task. We visualize the source spectral bases $\phi_{\mathcal{M}}$ (top) and the corresponding aligned target bases $\phi_{\mathcal{N}}$ (bottom). Due to the significant intrinsic distortion induced by partiality, the standard linear functional map $C_{\mathcal{NM}}$ fails to accurately align the spectral bases (highlighted in \textcolor{red}{red arrows}). In contrast, our neural functional map $h_{\theta_{\mathcal{NM}}}(\cdot)$ facilitates a more flexible, non-linear alignment (highlighted in \textcolor{ForestGreen}{green arrows}), resulting in superior matching accuracy and robustness under geometric irregularities.
    }
    \label{fig:alignment}
\end{figure}

Recent advancements in deep functional map frameworks have led to significant breakthroughs across various non-rigid 3D shape matching tasks. These include robust solutions for near-isometric~\cite{attaiki2023understanding} and non-isometric matching~\cite{donati2022deep, li2022learning}, as well as progress in multi-shape~\cite{cao2022unsupervised, eisenberger2023g, sun2023spatially}, multi-modal~\cite{cao2023self,jiang2023neural} and partial shape matching~\cite{attaiki2021dpfm, bracha2024wormhole, xie2025echomatch}. Notably, state-of-the-art frameworks~\cite{cao2023unsupervised, cao2024revisiting, bastian2024hybrid} have demonstrated remarkable versatility across diverse settings. However, despite these gains in standard scenarios, existing methods~\cite{cao2023unsupervised, cao2024revisiting} still fall short in more demanding conditions. Achieving accurate matching results remains a struggle when faced with significant partiality, topological noise, or the unstructured nature of raw point clouds. The failure of current functional map methods in these demanding scenarios is primarily rooted in a fundamental spectral misalignment: significant intrinsic distortion prevents the spectral bases of the two shapes from being accurately aligned via simple \emph{linear} transformations, see~\cref{fig:teaser}. Because the standard functional map framework~\cite{ovsjanikov2012functional} relies on the assumption that a linear operator can effectively map one basis to another, it becomes inherently ill-equipped to reconcile the non-linear basis shifts caused by partiality or topological noise as shown in~\cref{fig:alignment}. To overcome the aforementioned bottleneck, we draw inspiration from recent work on map refinement~\cite{vigano2025nam} and propose an unsupervised learning framework for \textit{non-linear neural functional maps} (NFM). 

In our approach we specifically substitute the standard linear functional map~\cite{ovsjanikov2012functional} with a non-linear neural functional map, which is modeled as an MLP, to facilitate more flexible spectral basis alignment. Unlike the previous refinement method~\cite{vigano2025nam} that requires accurate initial vertex-wise correspondences to optimize neural functional maps individually, we propose an unsupervised framework that learns a generalizable hyper-network using a training set {to learn priors over the training distribution}. At inference time, our hyper-network dynamically predicts the MLP weights conditioned on standard functional maps. To facilitate training, we introduce a novel spectral basis alignment loss, ensuring that the spectral bases are accurately aligned via the resulting non-linear transformation. Due to its design simplicity and flexibility, our framework can be seamlessly integrated into state-of-the-art deep functional map pipelines~\cite{cao2023self, cao2023unsupervised}. We demonstrate through extensive experiments that our method substantially improves matching accuracy in demanding scenarios.
Our primary contributions are summarized as follows:
\begin{itemize} 
\item We propose the first unsupervised learning framework for \textit{non-linear neural functional maps} (NFM), enabling flexible spectral basis alignment in challenging scenarios. 
\item We introduce a generalizable hyper-network that dynamically predicts the weights of these neural functional maps {conditioned on standard functional maps}, alongside a novel unsupervised spectral alignment loss to facilitate training. 
\item We demonstrate that our framework can be seamlessly integrated into existing state-of-the-art deep functional map pipelines, yielding substantial performance gains, particularly in the presence of partiality and topological noise. \end{itemize}
%
%
\section{Related Work} \label{sec:related_work}
Non-rigid 3D shape matching is an established research area in computer vision and graphics, with a rich history of development~\cite{van2011survey, tam2012registration}. Rather than providing an exhaustive literature review, we focus our discussion on contemporary approaches, specifically those built upon the functional map framework~\cite{ovsjanikov2012functional}.
%
\subsection{Deep Functional Map Methods}
 Rather than directly finding vertex-wise correspondences, which is often formulated as an expensive combinatorial optimization problem~\cite{windheuser2011geometrically, bernard2020mina, holzschuh2020simulated, roetzer2022scalable, gao2023sigma}, the functional map framework operates in the spectral domain. This paradigm encodes correspondence relationships into a compact matrix, termed the functional map, by projecting them onto a truncated set of basis functions~\cite{ovsjanikov2012functional} (typically the first $k$ Laplace-Beltrami eigenfunctions~\cite{levy2006laplace}). Due to its mathematical elegance and computational efficiency, the functional map framework has been extensively integrated into modern deep learning pipelines. Instead of using traditional handcrafted features~\cite{bronstein2010scale, aubry2011wave, salti2014shot}, deep learning approaches learn robust descriptors directly from training data to establish correspondences. Deep functional map methods are generally categorized as either supervised or unsupervised, depending on whether they require ground-truth correspondences for training. Supervised methods leverage ground-truth correspondences to optimize feature descriptors, primarily differing in their loss domain. While some minimize vertex-wise mismatches in the spatial domain~\cite{litany2017deep}, others operate on the spectral domain by regularizing the functional map itself~\cite{donati2020deep,attaiki2021dpfm}. More recent hybrid approaches~\cite{li2022srfeat} combine both constraints to achieve superior robustness. Unsupervised methods typically rely on the near-isometric deformation assumption to optimize correspondences without ground truth. Early approaches~\cite{halimi2019unsupervised, aygun2020heatkernel} penalize spatial mismatches by enforcing geodesic distance preservation. Subsequent works~\cite{roufosse2019unsupervised, sharma2020weakly} shift focus to the spectral domain, directly regularizing the structural properties (e.g., orthogonality and commutativity) of the functional map. Most recently, state-of-the-art performance has been achieved by frameworks~\cite{cao2023unsupervised, attaiki2023understanding, sun2023spatially} that couple functional maps with vertex-wise correspondences~\cite{ren2021discrete}. Despite these advances, the large intrinsic distortions inherent in partial or topologically noisy shapes prevent simple linear transformations from accurately aligning spectral bases. This fundamental limitation hinders the performance of current functional map methods in such demanding scenarios. To overcome this, we propose a novel unsupervised framework that learns \textit{non-linear neural functional maps}, enabling a more flexible and robust alignment of bases.

\subsection{Functional Map Refinement Methods}
A common strategy to enhance matching accuracy is to iteratively refine functional maps as a post-processing step. The foundational  ICP method iteratively converts functional maps to vertex-wise mappings via nearest-neighbor search~\cite{ovsjanikov2012functional}. Subsequent methods~\cite{rodola2015point, rodola2017regularized, ezuz2017deblurring} optimize this relationship by minimizing specific energy functions. For instance, PMF~\cite{vestner2017product} enforces bijectivity as a hard constraint, while ZoomOut~\cite{melzi2019zoomout} progressively increases spectral resolution during conversion. Other techniques focus on map properties: RHM~\cite{ezuz2019reversible} promotes reversible harmonic maps to reduce conformal distortion, and FSF~\cite{pai2021fast} utilizes theoretical spectral alignment for vertex-wise map recovery. For shape collections, another line of work leverages cycle-consistency~\cite{wang2013image, huang2014functional, bernard2019hippi, huang2020consistent,gao2021isometric}. The method most closely related to ours is NAM~\cite{vigano2025nam}, which replaces standard linear functional maps with neural adjoint maps (NAM). By extending ZoomOut~\cite{melzi2019zoomout} into a Neural ZoomOut framework, it achieves superior refinement, particularly for non-isometric shapes. However, as a two-stage "match-and-refine" pipeline, its performance remains heavily dependent on the quality of initial correspondences; erroneous starts often lead to sub-optimal local minima. In contrast, our approach introduces a generalizable hyper-network that directly predicts the neural functional map weights for any given shape pair. This shift from per-pair optimization to a predictive learning framework and thus avoids the need for careful initialization and eliminates the time-consuming optimization required by traditional refinement techniques.
%
%
\section{Background: Unsupervised Deep Functional Maps}
 \label{sec:background}\label{subsec:deep_fmap}
%
%
%
In this section, we provide the background for the contemporary unsupervised deep functional map framework~\cite{cao2023unsupervised, cao2024revisiting} and a more comprehensive discussion is provided in the supplementary materials.

Consider a pair of 3D shapes, $\mathcal{M}$ and $\mathcal{N}$, represented as triangle meshes with $n_{\mathcal{M}}$ and $n_{\mathcal{N}}$ vertices, respectively. We compute the associated positive semi-definite Laplace-Beltrami operators $L_{\mathcal{M}} \in \mathbb{R}^{n_{\mathcal{M}} \times n_{\mathcal{M}}}$ and $L_{\mathcal{N}} \in \mathbb{R}^{n_{\mathcal{N}} \times n_{\mathcal{N}}}$ using the standard cotangent discretization $L_{\mathcal{M}} = A_{\mathcal{M}}^{-1}W_{\mathcal{M}},$ where $A_{\mathcal{M}}$ denotes the diagonal matrix of lumped vertex areas and $W_{\mathcal{M}}$ represents the cotangent weight matrix~\cite{pinkall1993computing}. The spectral bases for each shape are then defined by the $k$ most dominant eigenfunctions, $\Phi_{\mathcal{M}} \in \mathbb{R}^{n_{\mathcal{M}} \times k}$ and $\Phi_{\mathcal{N}} \in \mathbb{R}^{n_{\mathcal{N}} \times k}$, obtained via the generalized eigenvalue decomposition of the respective Laplacians. A Siamese feature extractor (e.g., DiffusionNet~\cite{sharp2020diffusionnet}) processes both shapes to extract $c$-dimensional vertex-wise features, denoted as $F_{\mathcal{M}} \in \mathbb{R}^{n_{\mathcal{M}} \times c}$ and $F_{\mathcal{N}} \in \mathbb{R}^{n_{\mathcal{N}} \times c}$. Given the extracted features, the functional map $C_{\mathcal{MN}} \in \mathbb{R}^{k \times k}$ (relating the two spectral bases) is computed by solving a least-squares optimization problem 
\begin{equation}
            \label{eq:fmap}  C_{\mathcal{MN}}=\mathrm{argmin}_{C}~ E_{\mathrm{data}}\left(C\right)+\lambda E_{\mathrm{reg}}\left(C\right).
\end{equation}
Here, minimizing $E_{\mathrm{data}}=\left\|C\Phi_{\mathcal{M}}^{\dagger}F_{\mathcal{M}}-\Phi_{\mathcal{N}}^{\dagger}F_{\mathcal{N}}\right\|^{2}_{F}$ enforces descriptor preservation, while minimizing the regularization term $E_{\mathrm{reg}}$ imposes some form of structural properties (e.g., commutativity~\cite{ovsjanikov2012functional}). The operator $^{\dagger}$ denotes the Moore-Penrose inverse, which projects the features onto the spectral bases. Meanwhile, soft vertex-wise correspondences can be obtained via feature similarity
\begin{equation}
    \label{eq:soft_corr}
    \Pi_{\mathcal{NM}} = \mathrm{Softmax}\left( {F_{\mathcal{N}}F_{\mathcal{M}}^{T}} / \tau\right),
\end{equation}
where $\tau$ is the scaling factor to determine the softness of the correspondence matrix. During training, structural regularization (e.g.\ orthogonality, bijectivity~\cite{roufosse2019unsupervised}) is imposed on the computed functional maps $C_{\mathcal{MN}}$.
Moreover, to encourage that the computed functional map is associated with a valid vertex-wise map, an additional coupling loss is applied between the computed functional map and soft vertex-wise correspondences, i.e.,
\begin{equation}
    \label{eq:couple}
    L_{\mathrm{couple}} = \left\|C_{\mathcal{MN}} - \phi_{\mathcal{N}}^{\dagger}\Pi_{\mathcal{NM}}\phi_{\mathcal{M}}\right\|^{2}_{F}.
\end{equation}
During inference, the final vertex-wise correspondences can be obtained based on feature similarities using
    \begin{equation}
        \label{eq:nn_search}
        \Pi_{\mathcal{NM}} = \mathrm{NN}\left(F_{\mathcal{N}}, F_{\mathcal{M}}\right),
    \end{equation}
where $\mathrm{NN}$ denotes nearest neighbor search in $F_{\mathcal{N}}$ for each entry in $F_{\mathcal{M}}$. 

Despite its mathematical elegance, the standard deep functional map framework remains limited by its reliance on \emph{linear} transformations. This linear dependency proves insufficient when handling the significant intrinsic distortions present in partial or topologically noisy shapes, as a single matrix $C$ cannot accurately model the complex spectral shifts induced by such irregularities. This limitation motivates the proposal of \textit{non-linear neural functional maps}, which facilitate a more flexible and robust alignment of spectral bases. By transcending the limitations of linear transformations, it effectively accommodates the complex intrinsic distortions that standard functional maps fail to resolve.
%
%
\section{Our Unsupervised Neural Functional Map Method} \label{sec:method}
\begin{figure}[bht!]
    \centering
    \includegraphics[width=\textwidth]{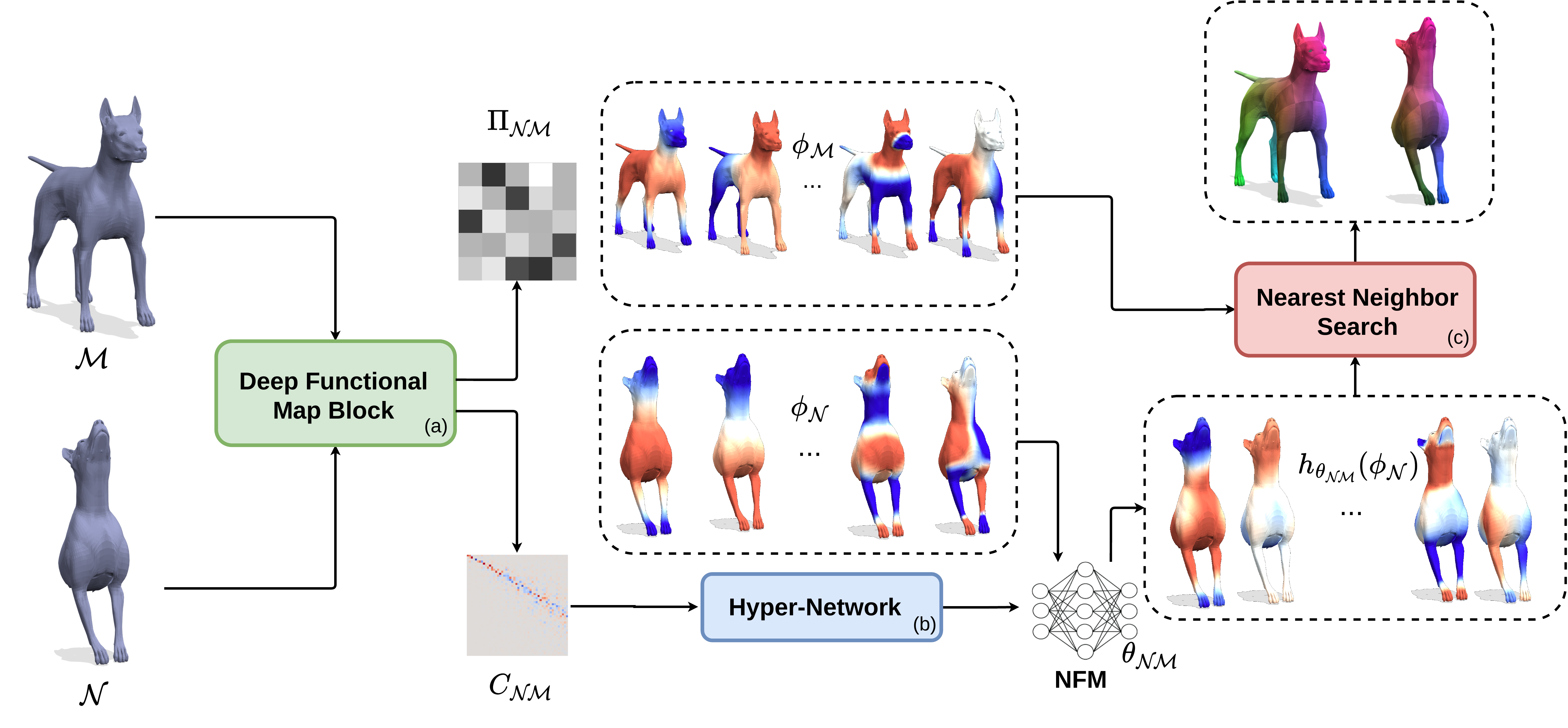}
    \caption{Our pipeline for unsupervised learning of a hyper-network to predict neural functional maps. Given a pair of 3D shapes, we first employ a standard deep functional map block to estimate an initial functional map and its corresponding vertex-wise correspondences. A hyper-network then predicts the weights of a neural functional map (i.e., $\theta_{\mathcal{NM}}$), conditioned on the initial functional map (i.e., $C_{\mathcal{NM}}$). This neural functional map facilitates a non-linear alignment of the target spectral bases with the source bases. Finally, refined vertex-wise correspondences are established via a nearest-neighbor search on this newly aligned spectral domain. Our unsupervised spectral alignment loss utilizes the initial vertex-wise correspondences $\Pi_{\mathcal{NM}}$ to train the hyper-network.
    }
    \label{fig:pipeline_nfm}
    \vspace{-5mm}
\end{figure}

We propose an unsupervised framework that replaces the traditional linear spectral transformation {under truncated bases} with non-linear neural functional maps {to compensate for this truncation-induced misalignment}. As illustrated in \cref{fig:pipeline_nfm}, our pipeline consists of three core stages: (a) an initial deep functional map block, (b) neural functional map prediction via a hyper-network, and (c) spectral alignment for final correspondence recovery.

First, we utilize a standard unsupervised functional map pipeline (\textit{cf.}~\cref{subsec:deep_fmap}) to estimate an initial functional map $C_{\mathcal{NM}}$ and soft vertex-wise correspondences $\Pi_{\mathcal{NM}}$. To overcome the inherent rigidity of this linear estimate, we introduce a hyper-network $H$ that takes $C_{\mathcal{NM}}$ to predict the parameters $\theta_{\mathcal{NM}}$ of a neural functional map $h_{\theta_{\mathcal{NM}}}$. This neural functional map (NFM) performs a non-linear alignment of the source spectral bases $\phi_{\mathcal{N}}$ to the target bases $\phi_{\mathcal{M}}$, effectively compensating for complex intrinsic distortions that a linear functional map cannot resolve. Finally, refined vertex-wise correspondences are established via a nearest-neighbor search in the resulting aligned spectral space. In the following sections, we detail the architectures of the hyper-network and neural functional map, alongside our unsupervised spectral alignment loss.

\subsection{Neural Functional Map}
Standard functional maps rely on a linear transformation matrix $C_{\mathcal{NM}} \in \mathbb{R}^{k \times k}$ to align the {truncated} spectral bases of two shapes. While computationally efficient, this linear formulation assumes that the relationship between spectral coefficients is strictly additive and proportional, which often fails under non-isometric deformations, topological noise or partiality. To address this, we define a neural functional map $h_{\theta_{\mathcal{NM}}}: \mathbb{R}^k \to \mathbb{R}^k$, a non-linear function parameterized by a neural network.

Instead of a global matrix multiplication, the mapping is defined as:
\begin{equation}
    \hat{\phi}_{\mathcal{N}}(x) = h_{\theta_{\mathcal{NM}}}(\phi_{\mathcal{N}}(x)), \forall x\in V_{\mathcal{N}},
\end{equation}
where $\phi_{\mathcal{N}}(x)$ represents the spectral coefficients of the vertex $x$ on the source shape $\mathcal{N}$. The network $h_{\theta_{\mathcal{NM}}}$ acts as a learnable transformation between spectral bases that can align the basis functions in a high-dimensional feature space. This formulation allows for a more flexible alignment, as the neural network can learn to ignore noisy spectral components or compensate for non-linear shifts in the {truncated} spectral bases.

In our implementation, $h_{\theta_{\mathcal{NM}}}$ is designed as a lightweight multi-layer perceptron (MLP) with residual connections. Specifically, we use a three layer MLP (similar to NAM~\cite{vigano2025nam}) {with a skip-connection of the standard FM $C_{\mathcal{NM}}$}, i.e.,
\begin{equation}
    h_{\theta_{\mathcal{NM}}}(\phi_{\mathcal{N}}(x)) = C_{\mathcal{NM}}\phi_{\mathcal{N}}(x) + W^3\sigma(W^2\sigma(W^1\phi_{\mathcal{N}}(x))),
    \label{eq:nfm}
\end{equation}
where $\sigma$ is a non-linear activation function (i.e., ReLU), $C_{\mathcal{NM}} \in \mathbb{R}^{k\times k}, W^1 \in \mathbb{R}^{k\times k}, W^2 \in \mathbb{R}^{k\times k}, W^3 \in \mathbb{R}^{k\times k}$ are linear transformations.

In contrast to NAM~\cite{vigano2025nam}, our parameters $\theta_{\mathcal{NM}}$ are not fixed during training but are instead dynamically generated by the hyper-network based on the initial standard functional map. By operating directly in the spectral domain, the neural functional map maintains the advantages of low-dimensional representation while gaining the expressiveness of deep neural architectures. By isolating the residual component in \cref{eq:nfm}, it becomes evident that the neural functional map recovers the standard linear transformation when the non-linear residuals are zero. Consequently, this formulation can be viewed as a generalized functional map framework: it preserves the structural benefits of classical spectral alignment while providing the necessary degrees of freedom to model the non-linear deformations that a fixed matrix $C_{\mathcal{NM}}$ cannot capture.

\subsection{Hyper-Network for Neural Functional Map Prediction}
To achieve a generalized matching framework that avoids expensive per-pair optimization, we introduce a hyper-network $H$ to dynamically generate the parameters of the neural functional map. Unlike traditional refinement methods~\cite{melzi2019zoomout,vigano2025nam} that treat map parameters as variables to be optimized at test-time, our hyper-network learns a mapping from the spectral domain representation (i.e., functional maps) to the parameter space of the MLP {by exploiting priors over the training distribution}.

The hyper-network takes the initial functional map $C_{\mathcal{NM}} \in \mathbb{R}^{k \times k}$, which encapsulates the global geometric relationship between the two shapes, as input. It then predicts the weights $\theta_{\mathcal{NM}}$ for the neural functional map $h_{\theta}$, i.e.,
\begin{equation}
    \theta_{\mathcal{NM}} = H(C_{\mathcal{NM}}),
\end{equation}
where $\theta_{\mathcal{NM}} = \{W^1_{\mathcal{NM}}, W^2_{\mathcal{NM}}, W^3_{\mathcal{NM}}\}
$.
By conditioning the neural mapping on $C_{\mathcal{NM}}$, the hyper-network allows the model to adapt its non-linear alignment strategy based on the specific distortion profile of the input pair characterized by the initial functional map {(e.g., slanted-diagonal structure under partiality)}.

Given that the initial functional map $C_{\mathcal{NM}}$ is a structured 2D representation of spectral correlations, we interpret it as a single-channel image as commonly done in prior functional map generation methods~\cite{zhuravlev2025denoising,pierson2025diffumatch}. We then employ a Vision Transformer (ViT) architecture~\cite{dosovitskiy2020image} for the hyper-network to exploit its self-attention mechanism, which is uniquely suited for capturing the global dependencies across different frequency bands within the functional map. Specifically, $H$ treats the $k \times k$ functional map as input and produces a $3 \times k \times k$ output. This output is interpreted as a three-channel feature map, where each channel corresponds to a specific linear layer for the neural functional map $h_{\theta}$ as defined in~\cref{eq:nfm}. By formulating the parameter prediction as an image-to-image translation task, the ViT can learn to recognize and correct complex patterns of spectral misalignment induced by partiality and topological changes, such as those appearing as off-diagonal noise or block-diagonal shifts~\cite{rodola2017partial}. 

\subsection{Unsupervised Spectral Alignment Loss}
To train the hyper-network without ground-truth supervision, we introduce a spectral alignment loss that encourages the neural functional map to find a non-linear mapping consistent with the estimated spatial correspondences. Let $\Pi_{\mathcal{NM}} \in [0, 1]^{n_{\mathcal{N}} \times n_{\mathcal{M}}}$ represent the soft vertex-wise correspondence matrix obtained from the initial deep functional map block (\textit{cf.}~\cref{eq:soft_corr}). This matrix represents the probability of a vertex on the target shape $\mathcal{M}$ being mapped to a vertex on the source shape $\mathcal{N}$.

The unsupervised spectral alignment loss is defined as the discrepancy between the target bases and the non-linearly transformed source bases in the spectral domain, i.e.,
\begin{equation}
    \label{eq:l_align}
    \mathcal{L}_{\mathrm{align}} = \left\|\Pi_{\mathcal{NM}} \phi_{\mathcal{M}} - h_{\theta_{\mathcal{NM}}}(\phi_{\mathcal{N}}) \right\|_F^2.
\end{equation}
Here, the soft correspondence matrix $\Pi_{\mathcal{NM}}$ acts as a spatial bridge, transporting the per-vertex spectral signals from the target domain to the source domain. The neural functional map $h_{\theta_{\mathcal{NM}}}$ is then tasked with non-linearly warping the source bases to match this transported target bases. By minimizing this objective, the hyper-network learns to predict parameters $\theta_{\mathcal{NM}}$ that correct for the misalignment presents in the initial linear functional map.

\section{Experimental Results}
\label{sec:experiments}
In this section, we evaluate the performance of our unsupervised neural functional map framework against state-of-the-art baselines across several challenging 3D shape matching scenarios, including topological noise, partiality, and multi-modal data. To demonstrate the versatility and robustness of our approach, we integrate our non-linear neural functional map learning module into two leading unsupervised frameworks: the state-of-the-art deep functional map pipeline~\cite{cao2023unsupervised} for topologically noisy and partial shape matching, and the self-supervised functional map learning framework~\cite{cao2023self} for multi-modal matching between triangle meshes and point clouds. By maintaining consistent experimental settings with these baselines, we highlight the specific performance gains obtained by our unsupervised neural functional map learning framework.
\subsection{Matching with Topological Noise}
\begin{figure}[!bht]
    \begin{tabular}{cc}

    \hspace{-1cm}
    \resizebox{0.40\linewidth}{!}{
\def\filename{kid00-kid01}
\def\pathOurs{figs/ours/topkids/}
\def\pathURSSM{figs/ulrssm/topkids/}
\def\pathAttn{figs/attnfmaps/topkids/}
\def\hspaceCols{0cm}
\def\wspaceRows{0cm}
\def\height{3.8cm}
\def\width{3.4cm}
\def\heightT{\height}
\def\widthT{\width}
\begin{tabular}{cc}%
    \setlength{\tabcolsep}{0pt} 
    {\scriptsize Source} & {\scriptsize AttnFMaps} \\
    \vspace{\wspaceRows}
    \hspace{\hspaceCols}
    \includegraphics[height=\heightT, width=\widthT]{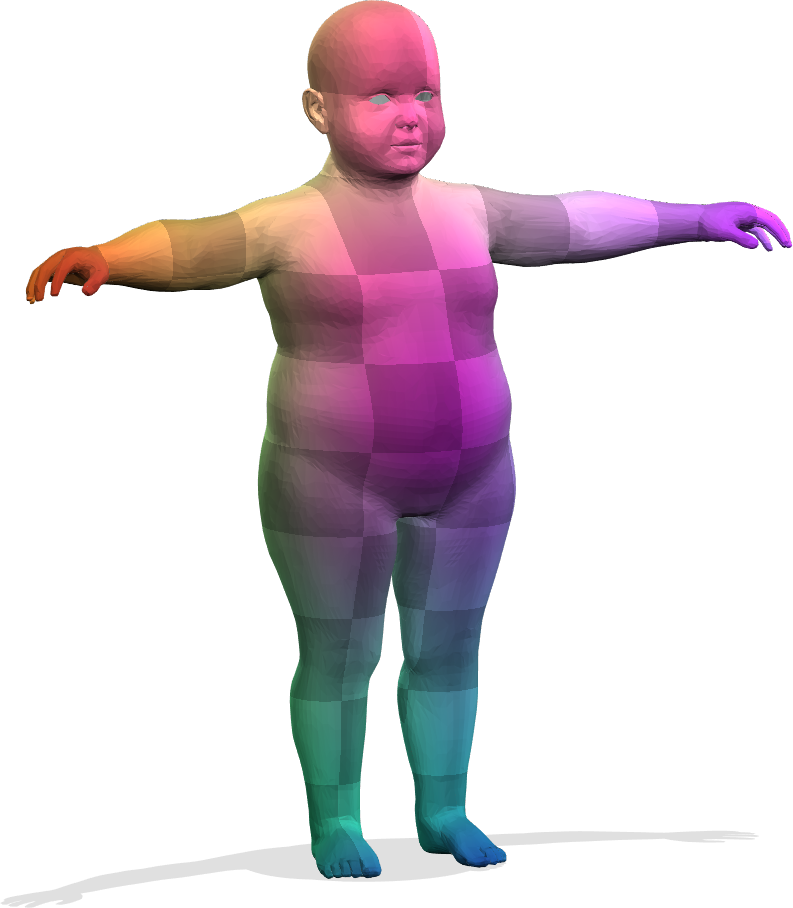}&
    \hspace{\hspaceCols}
    \includegraphics[height=\heightT, width=\widthT]{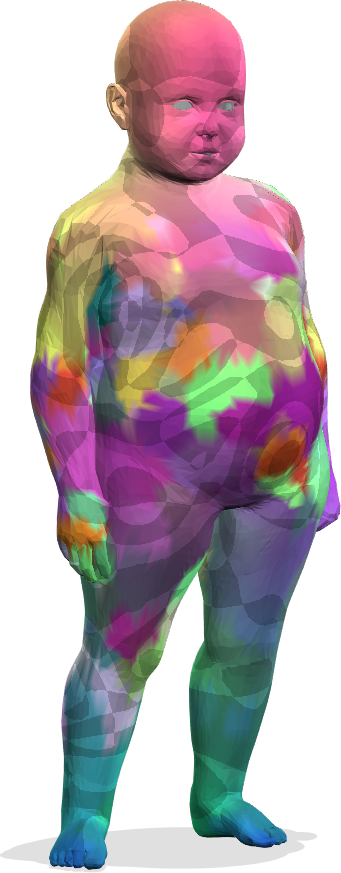}
    \\
    {\scriptsize ULRSSM} & {\scriptsize Ours} \\
    \vspace{\wspaceRows}
    \hspace{\hspaceCols}
    \begin{overpic}[height=\heightT, width=\widthT]{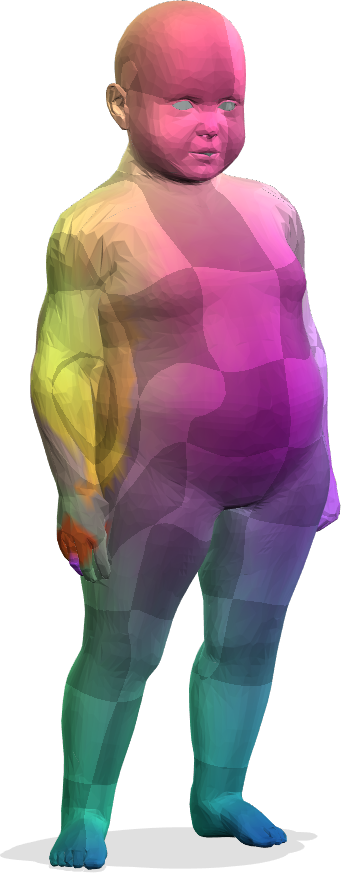}
       \put(6,37){\includegraphics[height=0.8cm]{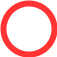}}
    \end{overpic}
    &
    \hspace{\hspaceCols}
    \begin{overpic}[height=\heightT, width=\widthT]{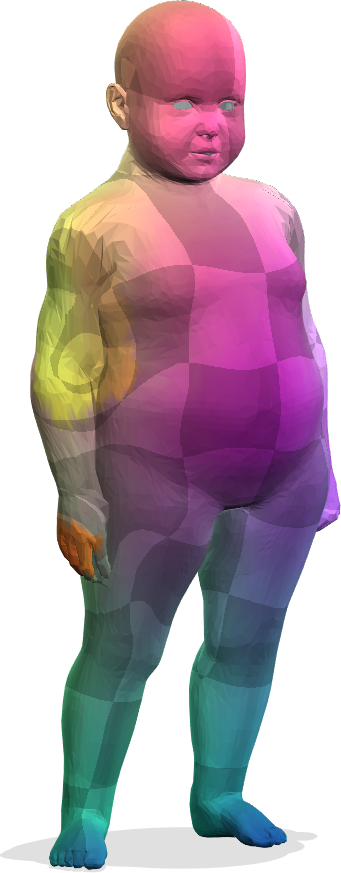}
       \put(6,37){\includegraphics[height=0.8cm]{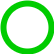}}
    \end{overpic}
\end{tabular}} 
    &
    \hspace{0.5cm}
    \resizebox{0.60\linewidth}{!}{
    \setlength{\tabcolsep}{4.5pt}
    \small
    \begin{tabular}{@{}lcc@{}}
    \toprule
    \multicolumn{1}{c}{\textbf{Geo. error ($\times$100)}}      & \multicolumn{1}{c}{\textbf{TOPKIDS}} & \multicolumn{1}{c}{\textbf{Intrinsic/Extrinsic}}
    \\ \midrule
    \multicolumn{3}{c}{Axiomatic Methods} \\
    \multicolumn{1}{l}{ZoomOut~\cite{melzi2019zoomout}}  & \multicolumn{1}{c}{33.7} & \multicolumn{1}{c}{Intrinsic}  \\ 
    \multicolumn{1}{l}{Smooth Shells~\cite{eisenberger2020smooth}}  & \multicolumn{1}{c}{11.8} & \multicolumn{1}{c}{Both}\\
    \multicolumn{1}{l}{DiscreteOp~\cite{ren2021discrete}}  & \multicolumn{1}{c}{35.5} & \multicolumn{1}{c}{Intrinsic}
    \\\midrule
    \multicolumn{3}{c}{Unsupervised Methods} \\
    \multicolumn{1}{l}{UnsupFMNet~\cite{halimi2019unsupervised}}  & \multicolumn{1}{c}{38.5} & \multicolumn{1}{c}{Intrinsic}\\
    \multicolumn{1}{l}{SURFMNet~\cite{roufosse2019unsupervised}}  & \multicolumn{1}{c}{48.6} & \multicolumn{1}{c}{Intrinsic}\\
    \multicolumn{1}{l}{WSupFMNet~\cite{sharma2020weakly}}  & \multicolumn{1}{c}{47.9} & \multicolumn{1}{c}{Intrinsic}\\
    \multicolumn{1}{l}{Deep Shells~\cite{eisenberger2020deep}}  & \multicolumn{1}{c}{13.7} & \multicolumn{1}{c}{Both} \\
    \multicolumn{1}{l}{NeuroMorph~\cite{eisenberger2021neuromorph}}  & \multicolumn{1}{c}{13.8} & \multicolumn{1}{c}{Both} \\
    \multicolumn{1}{l}{AttnFMaps~\cite{li2022learning}}  & \multicolumn{1}{c}{23.4} & \multicolumn{1}{c}{Intrinsic} \\
    \multicolumn{1}{l}{AttnFMaps-Fast~\cite{li2022learning}}  & \multicolumn{1}{c}{28.5} & \multicolumn{1}{c}{Intrinsic}\\
    \multicolumn{1}{l}{ULRSSM~\cite{cao2023unsupervised}}  & \multicolumn{1}{c}{{9.2}} & \multicolumn{1}{c}{Intrinsic} \\
    \multicolumn{1}{l}{Ours}  & \multicolumn{1}{c}{\textbf{6.7}} & \multicolumn{1}{c}{Intrinsic} \\
    \hline
    \end{tabular}
} \\
    \end{tabular}
    \vspace{-2mm}
    \caption{\textbf{Results on the TOPKIDS dataset.} We categorize existing approaches into fully intrinsic methods (based strictly on functional maps) and extrinsic-aware methods that utilize additional spatial information (e.g., rigid alignment). Our method achieves a substantial performance margin over all baselines, notably even outperforming even those methods that rely on extrinsic priors.
    }
    \label{fig:topkids}
\vspace{-5mm}
\end{figure}

\textbf{Datasets.} Real-world 3D scans often exhibit significant topological inconsistencies, such as self-intersections of distinct surfaces. These irregularities pose a severe challenge for existing matching methods: they distort the intrinsic geometry, causing spectral shifts that compromise functional map methods~\cite{lahner2016shrec}, and they introduce non-manifold artifacts that disrupt spatial registration approaches~\cite{eisenberger2023g}. To evaluate the robustness of our neural functional map method in the presence of such noise, we conduct experiments on the \textbf{TOPKIDS} dataset~\cite{lahner2016shrec}, which contains non-rigid shapes with varying levels of topological changes. Here the shapes with topological noise are expected to be matched to a clean template shape with T pose, as shown in~\cref{fig:topkids} left.

\noindent \textbf{Results.} The results on the \textbf{TOPKIDS} dataset demonstrate the superior robustness of our neural functional map framework. We use the mean geodesic error~\cite{kim2011blended} as our evaluation metric. As shown in~\cref{fig:topkids}, our method achieves a significant reduction in mean geodesic error compared to both functional map methods and methods based on spatial registration. Notably, our approach maintains high correspondence accuracy even in cases of severe self-intersection where standard linear methods suffer from failure. The performance gain can be attributed to the non-linear flexibility of the neural functional map. In the presence of topological noise, the spectral bases undergo localized shifts that cannot be modeled by a linear matrix (i.e., functional map). In contrast, our neural functional map learns to warp the spectral domain non-linearly. By effectively re-aligning these distorted basis functions, the neural functional map preserves the underlying geometric structure, allowing for accurate pointwise recovery where linear methods typically fail.

\subsection{Partial Shape Matching}
\textbf{Datasets.} For partial shape matching, we use the \textbf{\shrec{}} dataset~\cite{cosmo2016shrec}, which is based on the synthetic shapes of \tosca{}~\cite{bronstein2008numerical}. It contains $76$ nearly-isometric shapes from eight different classes of humans and animals. Each class has one template mesh in a neutral pose.
The partial shapes are divided into \cuts{} and \holes{}.
The \cuts{} dataset contains partial shapes cut by a plane, whereas the \holes{} dataset contains shapes with irregular holes that are produced by an erosion process. We rely on the \cutsTwentyFour{} split~\cite{ehm_partial--partial_2024} to ensure that no shapes from the test set are used for training. We also test on the \textbf{\pfaust{}} dataset~\cite{bracha2023partial}, a modification of the FAUST dataset~\cite{bogo2014faust}, where $10$ different humans ($8$ are used for training and $2$ for testing) are seen in $10$ different poses. \pfaust{} processes the dataset by adding artificial holes of varying difficulty, including \pfaustM{} (medium difficulty) and \pfaustH{} (hard difficulty). Following recent advancements in hybrid geometric-visual descriptors~\cite{dutt2024diffusion, wang2025kh}, we incorporate DINOv3 features~\cite{simeoni2025dinov3} to further improve matching performance. Specifically, we extract high-level semantic features from multi-view rendered images of each shape and lift them back to the shape vertices~\cite{dutt2024diffusion}. This allows our model to leverage robust visual priors alongside geometric information. To ensure a fair and transparent evaluation, we report matching accuracy using two distinct feature configurations: first, a version with pure vertex coordinates to maintain consistency with prior literature~\cite{cao2023unsupervised,attaiki2021dpfm}; second, a version enhanced by DINOv3 visual descriptors. In this way, it allows us to isolate the gains of our neural functional map from the improvements attributed to high-level semantic feature representations.

\begin{figure}[bht!]
\begin{tabular}{cc}
  \resizebox{0.60\linewidth}{!}{
    \setlength{\tabcolsep}{4.5pt}
    \small
 \begin{tabular}{@{}lcccc@{}}
    \toprule
    \multicolumn{1}{c}{\textbf{Geo. error ($\times$100)}}      & \multicolumn{2}{c}{\textbf{SHREC16}} &  \multicolumn{2}{c}{\textbf{PFAUST}}
    \\ 
    \cmidrule(lr){2-3} \cmidrule(lr){4-5}
    \multicolumn{1}{l}{} & \multicolumn{1}{c}{\textbf{CUTS}} & \multicolumn{1}{c}{\textbf{HOLES}} & \multicolumn{1}{c}{\textbf{Medium}} & \multicolumn{1}{c}{\textbf{Hard}} \\
    \midrule
    \multicolumn{5}{c}{Supervised Methods} \\
    \multicolumn{1}{l}{GeomFMaps~\cite{donati2020deep}}  & \multicolumn{1}{c}{12.8} & \multicolumn{1}{c}{15.3} & \multicolumn{1}{c}{4.3} & \multicolumn{1}{c}{6.6}  \\ 
    \multicolumn{1}{l}{\textit{+ DINOv3}}  & \multicolumn{1}{c}{3.2} & \multicolumn{1}{c}{6.8} & \multicolumn{1}{c}{2.8} & \multicolumn{1}{c}{3.8}  \\ 
    \multicolumn{1}{l}{DPFM-sup~\cite{attaiki2021dpfm}}  & \multicolumn{1}{c}{3.2} & \multicolumn{1}{c}{13.1} & \multicolumn{1}{c}{3.0} & \multicolumn{1}{c}{6.8} \\ 
    \multicolumn{1}{l}{\textit{+ DINOv3}}  & \multicolumn{1}{c}{3.2} & \multicolumn{1}{c}{6.6} & \multicolumn{1}{c}{2.8} & \multicolumn{1}{c}{3.8}  \\ 
    \midrule
    \multicolumn{5}{c}{Unsupervised Methods} \\
    \multicolumn{1}{l}{DPFM-unsup~\cite{attaiki2021dpfm}}  & \multicolumn{1}{c}{9.0} & \multicolumn{1}{c}{20.5} & \multicolumn{1}{c}{9.3} & \multicolumn{1}{c}{12.7}\\
    \multicolumn{1}{l}{\textit{+ DINOv3}}  & \multicolumn{1}{c}{10.8} & \multicolumn{1}{c}{10.4} & \multicolumn{1}{c}{3.8} & \multicolumn{1}{c}{6.5}   \\ 
    \multicolumn{1}{l}{ULRSSM~\cite{cao2023unsupervised}}  & \multicolumn{1}{c}{3.2} & \multicolumn{1}{c}{8.2} & \multicolumn{1}{c}{7.5} & \multicolumn{1}{c}{9.2} \\
    \multicolumn{1}{l}{\textit{+ DINOv3}}  & \multicolumn{1}{c}{4.3} & \multicolumn{1}{c}{9.7} & \multicolumn{1}{c}{\textbf{1.7}} & \multicolumn{1}{c}{5.2}  \\ 
    \multicolumn{1}{l}{Ours}  & \multicolumn{1}{c}{2.8} & \multicolumn{1}{c}{7.6} & \multicolumn{1}{c}{7.0} & \multicolumn{1}{c}{7.9} \\
    \multicolumn{1}{l}{\textit{+ DINOv3}}  & \multicolumn{1}{c}{\textbf{1.9}} & \multicolumn{1}{c}{\textbf{4.7}} & \multicolumn{1}{c}{\textbf{1.7}} & \multicolumn{1}{c}{\textbf{2.3}}   \\ 
    \hline
    \end{tabular}
  }
    &  
     \hspace{-0.8cm}
    \resizebox{0.40\linewidth}{!}{
\def\filename{cat-cuts_cat_shape_17}
\def\pathOurs{figs/ours/cuts/}
\def\pathURSSM{figs/ulrssm/cuts/}
\def\pathDPFM{figs/dpfm/cuts/}
\def\hspaceCols{0.2cm}
\def\wspaceRows{0cm}
\def\height{3.0cm}
\def\width{2.5cm}
\def\heightT{\height}
\def\widthT{\width}
\def\heightM{2.5cm}
\def\widthM{2.0cm}
\begin{tabular}{cc}%
    \setlength{\tabcolsep}{0pt} 
    {\small Source} & {\small DPFM-unsup} \\
    \vspace{\wspaceRows}
    \hspace{\hspaceCols}
    \includegraphics[height=\heightT, width=\widthT]{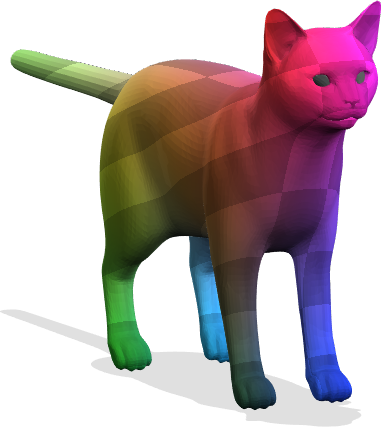}&
    \hspace{\hspaceCols}
    \includegraphics[height=\heightM, width=\widthM]{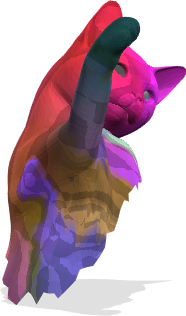}
    \\
    {\small ULRSSM} & {\small Ours} \\
    \vspace{\wspaceRows}
    \hspace{\hspaceCols}
    \includegraphics[height=\heightM, width=\widthM]{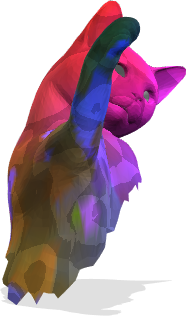}
    &
    \hspace{\hspaceCols}
    \includegraphics[height=\heightM, width=\widthM]{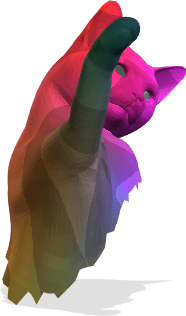}
\end{tabular}} 
\end{tabular}
    
    \caption{\textbf{Results on partial shape matching benchmarks.} Our method demonstrates superior robustness across diverse partiality patterns, significantly outperforming all unsupervised baselines in most cases. Notably, on the most challenging benchmarks (e.g., SHREC16 HOLES and PFAUST Hard), our unsupervised approach even significantly exceeds the performance of state-of-the-art supervised methods. This highlights the efficacy of our non-linear neural functional map, which accurately aligns spectral bases despite severe boundary artifacts and surface incompleteness.}
    \vspace{-5mm}
    \label{fig:cuts}
\end{figure}

\begin{figure}[bht!]
    \centering
\def\hspaceCols{0.2cm}
\def\wspaceRows{0cm}
\def\height{2.8cm}
\def\width{2.3cm}
\def\heightT{\height}
\def\widthT{\width}
\def\pathOurs{figs/ours/holes/}
\def\filenameOne{dog-holes_dog_shape_6_M}
\def\filenameTwo{dog-holes_dog_shape_6_N}
\def\filenameThree{dog-holes_dog_shape_12_N}
\def\filenameFour{dog-holes_dog_shape_25_N}
\def\filenameFive{dog-holes_dog_shape_15_N}
\def\filenameSix{dog-holes_dog_shape_22_N}
\begin{tabular}{cccccc}%
    \setlength{\tabcolsep}{0pt} 
    {\scriptsize Source} \\
    \hspace{\hspaceCols}
    \includegraphics[height=\heightT, width=\widthT]{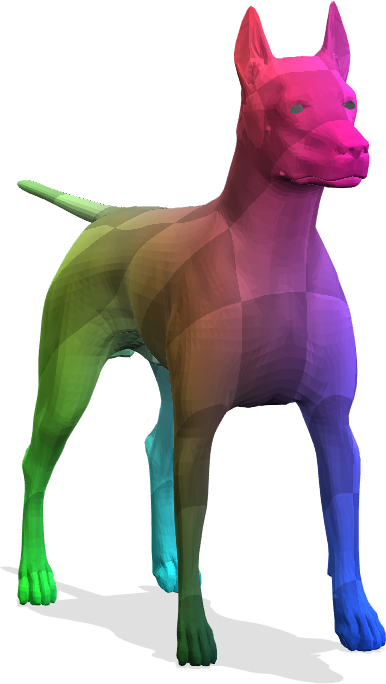}&
    \hspace{\hspaceCols}
    \includegraphics[height=\heightT, width=\widthT]{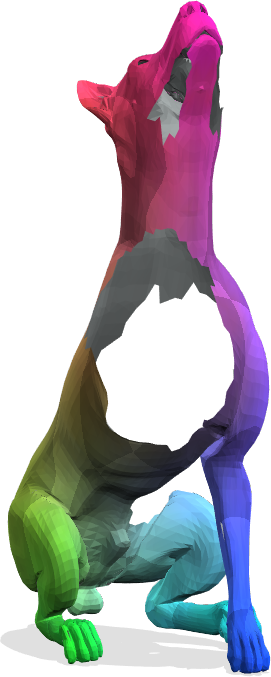}&
    \hspace{\hspaceCols}
    \includegraphics[height=\heightT, width=\widthT]{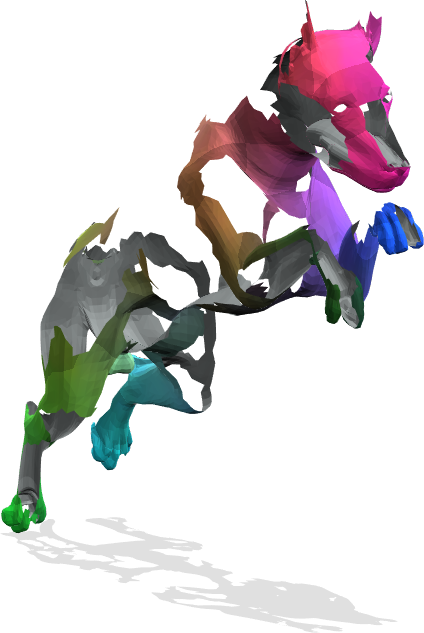}&
    \hspace{\hspaceCols}
    \includegraphics[height=\heightT, width=\widthT]{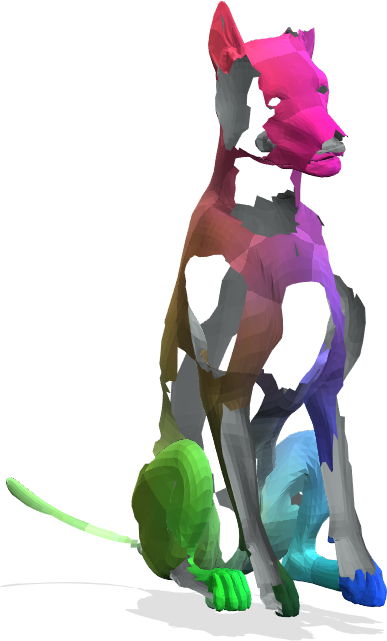}&
    \hspace{\hspaceCols}
    \includegraphics[height=\heightT, width=\widthT]{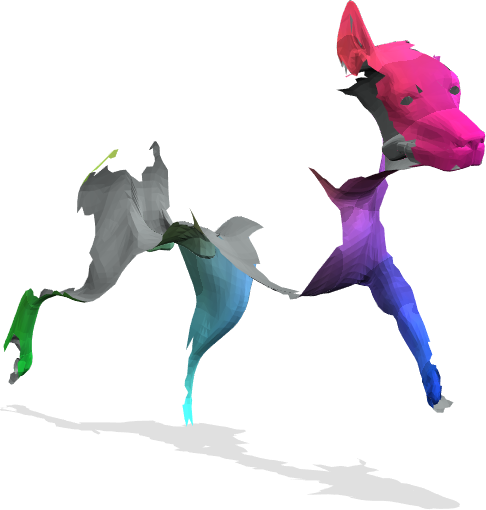}&
    \hspace{\hspaceCols}
    \includegraphics[height=\heightT, width=\widthT]{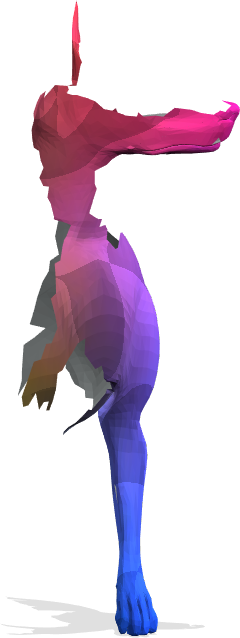}

\end{tabular}
    \caption{\textbf{Qualitative results on the SHREC16 HOLES benchmark.} Even under conditions of extreme partiality, our method recovers accurate, semantically consistent correspondences. This success is driven by the non-linear neural functional map, which provides the flexibility to align spectral bases that have been significantly distorted by the presence of large holes that traditional linear approximations typically collapse.}
    \label{fig:holes}
    \vspace{-6mm}
\end{figure}

\noindent \textbf{Results.} The results on partial shape matching benchmarks confirm that our method sets a new state-of-the-art for unsupervised partial shape matching. As indicated in~\cref{fig:cuts}, our method significantly outperforms all existing baseline methods in nearly every test scenario. Most notably, our unsupervised approach even surpasses the accuracy of established supervised methods on the most challenging benchmarks. This performance gap is particularly evident in datasets characterized by extreme missing geometry, where the inherent incompleteness of the shapes makes finding a stable basis alignment traditionally difficult. By enabling a flexible, non-linear transformation in the spectral domain based on our neural functional maps, our method can accurately correlate the relationship between partial and complete spectral bases, ensuring robust pointwise recovery even when a large portion of the shape is absent, as shown in~\cref{fig:holes}.

\subsection{Multi-Modal Shape Matching}
\label{subsec:multimodal}
\textbf{Datasets.} For multi-modal shape matching, we evaluate on the \textbf{FAUST}~\cite{bogo2014faust}, \textbf{SCAPE}~\cite{anguelov2005scape}, and \textbf{SHREC19}~\cite{melzi2019shrec} datasets. In alignment with prior works~\cite{cao2023self, jiang2023non,jiang2023neural}, we assess the model's ability to establish correspondences across different shape representations, specifically between triangle meshes and unstructured point clouds. The FAUST dataset consists of 100 shapes (10 people in 10 poses), where the evaluation is performed on the last 20 shapes. The SCAPE dataset contains 71 different poses of the same person, where the last 20 shapes are used for evaluation. The SHREC19 dataset is a more challenging benchmark dataset due to its significant variations in mesh connectivity and irregular sampling. It comprises 44 shapes and a total of 430 evaluation pairs. To be consistent with prior works~\cite{cao2023self, jiang2023non,jiang2023neural}, we use vertex coordinate as the input for all datasets.

\begin{table}[bht!]
    \centering
    \caption{\textbf{Results on multi-modal shape matching datasets.} We evaluate all methods
    on individual dataset for shapes represented as triangle meshes and point clouds. The first row indicates the training dataset and the second row indicates the test datasets. The numbers in parentheses indicates the mean geodesic error for point clouds. The best results in each column are \textbf{highlighted}.}
    \label{tab:multimodal}
     \resizebox{1.0\linewidth}{!}{
    \setlength{\tabcolsep}{4.5pt}
    \small
    \begin{tabular}{@{}lcccccc@{}}
    \toprule
    \multicolumn{1}{c}{\textbf{Geo. error ($\times$100)}}      & \multicolumn{3}{c}{\textbf{FAUST}} &  \multicolumn{3}{c}{\textbf{SCAPE}}
    \\ 
    \cmidrule(lr){2-4} \cmidrule(lr){5-7}
    \multicolumn{1}{l}{} & \multicolumn{1}{c}{\textbf{FAUST}} & \multicolumn{1}{c}{\textbf{SCAPE}} & \multicolumn{1}{c}{\textbf{SHREC19}} & \multicolumn{1}{c}{\textbf{SCAPE}} & \multicolumn{1}{c}{\textbf{FAUST}} & \multicolumn{1}{c}{\textbf{SHREC19}} \\
    \midrule
    \multicolumn{7}{c}{Supervised Methods} \\
    \multicolumn{1}{l}{3D-CODED~\cite{groueix20183d}}  & \multicolumn{1}{c}{2.5 (2.5)} & \multicolumn{1}{c}{31.0 (31.0)} & 
    \multicolumn{1}{c}{-} & \multicolumn{1}{c}{31.0 (31.0)} & \multicolumn{1}{c}{33.0 (33.0)} & 
    \multicolumn{1}{c}{-} \\ 
    \multicolumn{1}{l}{DiffFMaps~\cite{marin2020correspondence}}  & \multicolumn{1}{c}{3.6 (3.6)} & \multicolumn{1}{c}{19.0 (19.0)} & 
    \multicolumn{1}{c}{16.4 (16.4)} & \multicolumn{1}{c}{12.0 (12.0)} & \multicolumn{1}{c}{12.0 (12.0)} & 
    \multicolumn{1}{c}{17.6 (17.6)} \\ 
    \multicolumn{1}{l}{TransMatch~\cite{trappolini2021shape}} & \multicolumn{1}{c}{2.7 (2.7)} & \multicolumn{1}{c}{33.6 (33.6)} & 
    \multicolumn{1}{c}{21.0 (21.0)} & \multicolumn{1}{c}{18.6 (18.6)} & \multicolumn{1}{c}{18.3 (18.3)} & 
    \multicolumn{1}{c}{38.8 (38.8)} \\ 
    \midrule
    \multicolumn{7}{c}{Unsupervised/Self-Supervised Methods} \\
    \multicolumn{1}{l}{CorrNet3D~\cite{zeng2021corrnet3d}}   & \multicolumn{1}{c}{26.5 (26.5)} & \multicolumn{1}{c}{38.0 (38.0)} & 
    \multicolumn{1}{c}{44.0 (44.0)} & \multicolumn{1}{c}{37.3 (37.3)} & \multicolumn{1}{c}{36.2 (36.2)} & 
    \multicolumn{1}{c}{42.6 (42.6)} \\ 
    \multicolumn{1}{l}{DCP~\cite{lang2021dpc}}  & \multicolumn{1}{c}{11.1 (11.1)} & \multicolumn{1}{c}{17.5 (17.5)} & 
    \multicolumn{1}{c}{31.0 (31.0)} & \multicolumn{1}{c}{17.3 (17.3)} & \multicolumn{1}{c}{11.2 (11.2)} & 
    \multicolumn{1}{c}{28.7 (28.7)} \\ 
    \multicolumn{1}{l}{NIE~\cite{jiang2023neural}}   & \multicolumn{1}{c}{5.5 (5.5)} & \multicolumn{1}{c}{15.0 (15.0)} & 
    \multicolumn{1}{c}{15.1 (15.1)} & \multicolumn{1}{c}{11.0 (11.0)} & \multicolumn{1}{c}{8.7 (8.7)} & 
    \multicolumn{1}{c}{15.6 (15.6)} \\ 
    \multicolumn{1}{l}{DFR~\cite{jiang2023non}}   & \multicolumn{1}{c}{3.0 (3.0)} & \multicolumn{1}{c}{6.3 (6.3)} & 
    \multicolumn{1}{c}{5.9 ({5.9})} & \multicolumn{1}{c}{2.9 (2.9)} & \multicolumn{1}{c}{4.0 (4.0)} & 
    \multicolumn{1}{c}{4.8 (4.8)} \\ 
    \multicolumn{1}{l}{SSMSM~\cite{cao2023self}}  & \multicolumn{1}{c}{2.0 (2.4)} & \multicolumn{1}{c}{9.2 (11.0)} & \multicolumn{1}{c}{7.5 (9.0)} & \multicolumn{1}{c}{3.1 (4.1)} & \multicolumn{1}{c}{7.2 (8.5)} & 
    \multicolumn{1}{c}{6.8 (7.3)} \\
    \multicolumn{1}{l}{Ours}   & \multicolumn{1}{c}{\textbf{1.8} (\textbf{2.2})} & \multicolumn{1}{c}{\textbf{4.6} (\textbf{6.0})} & \multicolumn{1}{c}{\textbf{5.0} (\textbf{5.7})} & \multicolumn{1}{c}{\textbf{2.3} (\textbf{2.8})} & \multicolumn{1}{c}{\textbf{3.2} (\textbf{3.7})} & 
    \multicolumn{1}{c}{\textbf{4.3} (\textbf{4.7})} \\
    \hline
    \end{tabular}
    \vspace{-5mm}
    }
\end{table}

\noindent \textbf{Results.} Across all evaluated benchmarks, our framework sets the new state of the art, consistently surpassing both supervised and self-supervised baselines (\textit{cf.}~\cref{tab:multimodal}) and {shows remarkable cross-dataset generalization ability for both mesh and point cloud}. In the multi-modal setting, the neural functional map proves particularly effective at bridging the spectral gap between triangle meshes and point clouds, where discretization differences often lead to basis misalignment. By outperforming existing self-supervised and supervised methods on FAUST, SCAPE, and the challenging SHREC’19 datasets, our approach demonstrates superior generalization and robustness. These results validate that our non-linear refinement is not only more expressive than standard linear transformations but also more resilient to the varied sampling densities and connectivity artifacts inherent in multi-modal data.

\section{Ablation Study}
\label{sec:ablation}
\subsection{Component Analysis}
To evaluate the contribution of each component within our framework, we conduct a series of ablation experiments on the TOPKIDS dataset, as its significant topological noise provides a rigorous test for method robustness. Our investigation focuses on two primary aspects: the architecture of the hyper-network and the depth (capacity) of the neural functional map.

First, we evaluate the impact of the hyper-network by replacing our default Vision Transformer (ViT) backbone with CNN and MLP alternatives. This comparison demonstrates that the global receptive field and self-attention mechanism of the ViT are essential for capturing long-range spectral dependencies required to predict coherent map parameters. Second, we ablate the complexity of the neural functional map itself by varying the number of hidden layers in its MLP structure, seeking the optimal balance between expressivity and regularization.

\noindent\textbf{(a) Hyper-network as MLP.} We replace the ViT with a standard multi-layer perceptron. This setting lacks the spatial and structural priors of the ViT, treating the input functional map as a flattened vector, which limits its ability to recognize localized spectral patterns.

\noindent\textbf{(b) Hyper-network as CNN.} We utilize a convolutional neural network as the hyper-network. While this introduces inductive biases for local feature extraction, it lacks the global context provided by self-attention, which is crucial for understanding the overall functional map structure.

\noindent\textbf{(c) 1-hidden layer.} We restrict the neural functional map to a single hidden layer. This shallow configuration serves as a "near-linear" baseline to test if minimal non-linearity is sufficient to correct spectral distortions.

\noindent\textbf{(d) 3-hidden layers.} We increase the depth of the neural functional map to three hidden layers. This setting evaluates whether more non-linear capacity can capture the complex warping required for topologically noisy shapes, compared to our 2-hidden layers default setting.

\noindent\textbf{(e) 4-hidden layers.} We test a deeper four-layer configuration for the neural functional map. This allows us to observe if excessive depth leads to overfitting or if the added capacity further aids in basis alignment.
\begin{table}[bht!]
    \centering
    \scriptsize
    \vspace{-3mm}
    \caption{Qualitative results of our ablation study on TOPKIDS dataset. Compared to different settings, our default setting obtains the best matching performance.}
    \label{tab:ablation}
    \setlength{\tabcolsep}{1pt}
    \begin{tabular}{@{}lcccccc@{}}
    \toprule
    \textbf{TOPKIDS} & {(a) MLP} & (b) CNN  & (c) 1-hidden layer & (d) 3-hidden layers & (e) 4-hidden layers & (f) Ours \\ \midrule
    \textbf{Geo. error} & 10.8 & 9.6 & 10.3 & 8.6 & 8.9 & \textbf{6.7} \\ 
    \bottomrule
\end{tabular}
\vspace{-6mm}   
\end{table}

\noindent\textbf{Results.} As shown in~\cref{tab:ablation}, our default configuration (ViT hyper-network with a 2-hidden-layer neural functional map) achieves the lowest geodesic error. The significant performance gap between the ViT and the CNN/MLP backbones (settings (a) and (b)) highlights that global context is vital for parameter prediction. Furthermore, the results for settings (c), (d), and (e) suggest that a 2-layer MLP provides the ideal zone for expressivity; fewer layers lack the flexibility to model distortions, while deeper layers do not yield additional gains, likely due to the difficulty of optimizing highly non-linear spectral transformations.

\subsection{Well-Posedness of Our Method}
To evaluate whether our hyper-network can exploit priors over the training distribution to recover high-frequency non-linear residual from standard functional alone, we conduct the following ablative experiments, as shown in~\cref{table:ablationwellposedness}.

\noindent{\textbf{(a) Swap $C$:} swap original $C$ with another pair's $C$. The high errors indicate that the hyper-network \emph{does not}  learn a constant dataset-specific  correction.
}

\noindent{\textbf{(b) Rnd-init:} randomly initialized hyper-network weights with  per-pair test-time optimization. High errors indicate that per-pair test-time optimization does not work, confirming that the hyper-network learns a generalized mapping from linear FM to NFM from the \emph{collective information} in the training dataset.}

\noindent{\textbf{(c) Cross-dataset:} train on CUTS, test on HOLES (and vice-versa). The results show that our hyper-network 
does not only learn dataset-specific bias but  generalizes across datasets.}

\noindent{\textbf{(d) $C$-corruption:} degradation of $C$ by Gaussian noise $\sigma=0.1$ during inference on test datasets. The results demonstrate the robustness of our hyper-network and that it does not overfit to training data bias.}

\begin{table}[bht!]
    \centering
    \scriptsize
    \caption{{Qualitative results on SHREC16 dataset. Results demonstrate our method successfully learns generalizable priors from training data distribution.}}
    \label{tab:well_posedness}
    \setlength{\tabcolsep}{6pt}
    \begin{tabular}{lccccc}
    \toprule
    Geo. error ($\times 100$) & {Ours} & Swap $C$ & Rnd-init & Cross-dataset & $C$-corruption \\
    \midrule
    CUTS    & \textbf{1.9}              & 21.7    & 57.3    & 3.0  & 3.1 \\
    HOLES   & \textbf{4.7}             & 32.8    & 61.9    & 5.7 & 6.2 \\
    \bottomrule
    \end{tabular}\label{table:ablationwellposedness}
\vspace{-6mm}      
\end{table}

{
\noindent\textbf{Results.} Our method successfully learns generalizable priors from training data. By construction (\cref{eq:nfm}), our NFM reduces to $C$ when the residual vanishes and operates entirely in the spectral domain. Overall, our hyper-network's task is to enrich a coarse spectral alignment with a non-linear correction, conditioned on the distortion profile that $C$ already encodes. To this end, our method can effectively exploit priors over the training distribution and generalize to other datasets.}
\section{Limitation and Future Work}
\label{sec:limitation}
Despite its robustness, our framework is limited by its fundamental reliance on spectral alignment. In cases of extreme geometric distortion where Laplace-Beltrami eigenfunctions become significantly degraded, the non-linear mapping may fail to bridge the resulting spectral gap. Future work could address this, for example by extending the framework to multi-shape matching through the enforcement of cycle-consistency as regularization. Specifically, learning a set of virtual universal spectral bases to serve as a canonical reference domain would allow the hyper-network to align an entire collection to a shared latent space, ensuring globally consistent correspondences for advanced joint shape analysis.

\section{Conclusion}
\label{sec:conclusion}

In this paper, we introduced a novel framework for 3D shape matching that moves beyond the limitations of standard functional maps by incorporating non-linear neural functional maps. By utilizing a ViT-based hyper-network, our method dynamically predicts the parameters of a non-linear functional map based on the standard functional map, providing the flexibility required to align spectral bases in the presence of significant distortions. Our extensive experiments across diverse benchmarks, including topologically noisy, partial, and multi-modal datasets, demonstrate that our approach consistently sets the new state of the art. Notably, our unsupervised framework often surpasses the accuracy of fully supervised alternatives, particularly in challenging scenarios where standard linear approximations collapse. By bridging the gap between spectral processing and deep non-linear architectures, this work provides a robust and versatile foundation for high-quality correspondence for 3D shape analysis and beyond. Ultimately, this paradigm shift toward adaptive spectral alignment paves the way for more robust matching algorithm capable of handling the inherent geometric inconsistencies of real-world data in fields such as robotics, medical imaging and digital twins.
\section{Acknowledgment}
{This work is supported by the ERC starting grant no. 101160648 (Harmony).}
\clearpage

%
%
\bibliographystyle{splncs04}
\bibliography{main}

\begin{thebibliography}{10}
\providecommand{\url}[1]{\texttt{#1}}
\providecommand{\urlprefix}{URL }
\providecommand{\doi}[1]{https://doi.org/#1}

\bibitem{anguelov2005scape}
Anguelov, D., Srinivasan, P., Koller, D., Thrun, S., Rodgers, J., Davis, J.: Scape: shape completion and animation of people. In: ACM SIGGRAPH (2005)

\bibitem{attaiki2023understanding}
Attaiki, S., Ovsjanikov, M.: Understanding and improving features learned in deep functional maps. In: CVPR (2023)

\bibitem{attaiki2021dpfm}
Attaiki, S., Pai, G., Ovsjanikov, M.: Dpfm: Deep partial functional maps. In: International Conference on 3D Vision (3DV) (2021)

\bibitem{aubry2011wave}
Aubry, M., Schlickewei, U., Cremers, D.: The wave kernel signature: A quantum mechanical approach to shape analysis. In: ICCV (2011)

\bibitem{aygun2020heatkernel}
Ayg\"un, M., L\"ahner, Z., Cremers, D.: Unsupervised dense shape correspondence using heat kernels. In: Conference on 3D Vision {(3DV)} (2020)

\bibitem{bastian2024hybrid}
Bastian, L., Xie, Y., Navab, N., L{\"a}hner, Z.: Hybrid functional maps for crease-aware non-isometric shape matching. In: CVPR (2024)

\bibitem{bernard2020mina}
Bernard, F., Suri, Z.K., Theobalt, C.: Mina: Convex mixed-integer programming for non-rigid shape alignment. In: CVPR (2020)

\bibitem{bernard2019hippi}
Bernard, F., Thunberg, J., Swoboda, P., Theobalt, C.: Hippi: Higher-order projected power iterations for scalable multi-matching. In: ICCV (2019)

\bibitem{bogo2014faust}
Bogo, F., Romero, J., Loper, M., Black, M.J.: Faust: Dataset and evaluation for 3d mesh registration. In: CVPR (2014)

\bibitem{bracha2023partial}
Bracha, A., Dag{\`e}s, T., Kimmel, R.: On unsupervised partial shape correspondence. ACCV  (2024)

\bibitem{bracha2024wormhole}
Bracha, A., Dag{\`e}s, T., Kimmel, R.: Wormhole loss for partial shape matching. In: NIPS (2024)

\bibitem{bronstein2008numerical}
Bronstein, A., Bronstein, M., Kimmel, R.: Numerical Geometry of Non-Rigid Shapes. Springer Publishing Company, Incorporated, 1 edn. (2008)

\bibitem{bronstein2010scale}
Bronstein, M.M., Kokkinos, I.: Scale-invariant heat kernel signatures for non-rigid shape recognition. In: CVPR (2010)

\bibitem{cao2022unsupervised}
Cao, D., Bernard, F.: Unsupervised deep multi-shape matching. In: ECCV (2022)

\bibitem{cao2023self}
Cao, D., Bernard, F.: Self-supervised learning for multimodal non-rigid 3d shape matching. In: CVPR (2023)

\bibitem{cao2023unsupervised}
Cao, D., Roetzer, P., Bernard, F.: Unsupervised learning of robust spectral shape matching. ACM Transactions on Graphics (ToG)  (2023)

\bibitem{cao2024revisiting}
Cao, D., Roetzer, P., Bernard, F.: Revisiting map relations for unsupervised non-rigid shape matching. In: 2024 International Conference on 3D Vision (3DV). IEEE (2024)

\bibitem{cosmo2016shrec}
Cosmo, L., Rodola, E., Bronstein, M.M., Torsello, A., Cremers, D., Sahillioglu, Y.: Shrec’16: Partial matching of deformable shapes. Proc. 3DOR  \textbf{2}(9), ~12 (2016)

\bibitem{deng2022survey}
Deng, B., Yao, Y., Dyke, R.M., Zhang, J.: A survey of non-rigid 3d registration. In: Computer Graphics Forum. vol.~41, pp. 559--589. Wiley Online Library (2022)

\bibitem{dinh2005texture}
Dinh, H.Q., Yezzi, A., Turk, G.: Texture transfer during shape transformation. ACM Transactions on Graphics (ToG)  \textbf{24}(2),  289--310 (2005)

\bibitem{donati2022deep}
Donati, N., Corman, E., Ovsjanikov, M.: Deep orientation-aware functional maps: Tackling symmetry issues in shape matching. In: CVPR (2022)

\bibitem{donati2020deep}
Donati, N., Sharma, A., Ovsjanikov, M.: Deep geometric functional maps: Robust feature learning for shape correspondence. In: CVPR (2020)

\bibitem{dosovitskiy2020image}
Dosovitskiy, A.: An image is worth 16x16 words: Transformers for image recognition at scale. arXiv preprint arXiv:2010.11929  (2020)

\bibitem{dutt2024diffusion}
Dutt, N.S., Muralikrishnan, S., Mitra, N.J.: Diffusion 3d features (diff3f): Decorating untextured shapes with distilled semantic features. In: CVPR (2024)

\bibitem{egger20203d}
Egger, B., Smith, W.A., Tewari, A., Wuhrer, S., Zollhoefer, M., Beeler, T., Bernard, F., Bolkart, T., Kortylewski, A., Romdhani, S., et~al.: 3d morphable face models—past, present, and future. ACM Transactions on Graphics (ToG)  \textbf{39}(5),  1--38 (2020)

\bibitem{ehm_partial--partial_2024}
Ehm, V., Gao, M., Roetzer, P., Eisenberger, M., Cremers, D., Bernard, F.: Partial-to-partial shape matching with geometric consistency. In: CVPR (2024)

\bibitem{eisenberger2020smooth}
Eisenberger, M., Lahner, Z., Cremers, D.: Smooth shells: Multi-scale shape registration with functional maps. In: CVPR (2020)

\bibitem{eisenberger2021neuromorph}
Eisenberger, M., Novotny, D., Kerchenbaum, G., Labatut, P., Neverova, N., Cremers, D., Vedaldi, A.: Neuromorph: Unsupervised shape interpolation and correspondence in one go. In: CVPR (2021)

\bibitem{eisenberger2020deep}
Eisenberger, M., Toker, A., Leal-Taix{\'e}, L., Cremers, D.: Deep shells: Unsupervised shape correspondence with optimal transport. NIPS  (2020)

\bibitem{eisenberger2023g}
Eisenberger, M., Toker, A., Leal-Taix{\'e}, L., Cremers, D.: G-msm: Unsupervised multi-shape matching with graph-based affinity priors. In: CVPR (2023)

\bibitem{ezuz2017deblurring}
Ezuz, D., Ben-Chen, M.: Deblurring and denoising of maps between shapes. In: Computer Graphics Forum. Wiley Online Library (2017)

\bibitem{ezuz2019reversible}
Ezuz, D., Solomon, J., Ben-Chen, M.: Reversible harmonic maps between discrete surfaces. ACM Transactions on Graphics (ToG)  \textbf{38}(2),  1--12 (2019)

\bibitem{gao2021isometric}
Gao, M., Lahner, Z., Thunberg, J., Cremers, D., Bernard, F.: Isometric multi-shape matching. In: CVPR (2021)

\bibitem{gao2023sigma}
Gao, M., Roetzer, P., Eisenberger, M., L{\"a}hner, Z., Moeller, M., Cremers, D., Bernard, F.: Sigma: Scale-invariant global sparse shape matching. In: ICCV (2023)

\bibitem{groueix20183d}
Groueix, T., Fisher, M., Kim, V.G., Russell, B.C., Aubry, M.: 3d-coded: 3d correspondences by deep deformation. In: ECCV (2018)

\bibitem{halimi2019unsupervised}
Halimi, O., Litany, O., Rodola, E., Bronstein, A.M., Kimmel, R.: Unsupervised learning of dense shape correspondence. In: CVPR (2019)

\bibitem{holzschuh2020simulated}
Holzschuh, B., L{\"a}hner, Z., Cremers, D.: Simulated annealing for 3d shape correspondence. In: 2020 International Conference on 3D Vision (3DV) (2020)

\bibitem{huang2014functional}
Huang, Q., Wang, F., Guibas, L.: Functional map networks for analyzing and exploring large shape collections. ACM Transactions on Graphics (ToG)  \textbf{33}(4),  1--11 (2014)

\bibitem{huang2020consistent}
Huang, R., Ren, J., Wonka, P., Ovsjanikov, M.: Consistent zoomout: Efficient spectral map synchronization. In: Computer Graphics Forum. Wiley Online Library (2020)

\bibitem{jiang2023neural}
Jiang, P., Sun, M., Huang, R.: Neural intrinsic embedding for non-rigid point cloud matching. In: CVPR (2023)

\bibitem{jiang2023non}
Jiang, P., Sun, M., Huang, R.: Non-rigid shape registration via deep functional maps prior. In: NIPS (2023)

\bibitem{kim2011blended}
Kim, V.G., Lipman, Y., Funkhouser, T.: Blended intrinsic maps. ACM Transactions on Graphics (ToG)  \textbf{30}(4),  1--12 (2011)

\bibitem{lahner2016shrec}
L{\"a}hner, Z., Rodola, E., Bronstein, M.M., Cremers, D., Burghard, O., Cosmo, L., Dieckmann, A., Klein, R., Sahillioglu, Y.: Shrec’16: Matching of deformable shapes with topological noise. Proc. 3DOR  \textbf{2}(10.2312) (2016)

\bibitem{lang2021dpc}
Lang, I., Ginzburg, D., Avidan, S., Raviv, D.: Dpc: Unsupervised deep point correspondence via cross and self construction. In: International Conference on 3D Vision (3DV). IEEE (2021)

\bibitem{levy2006laplace}
L{\'e}vy, B.: Laplace-beltrami eigenfunctions towards an algorithm that" understands" geometry. In: IEEE International Conference on Shape Modeling and Applications 2006 (SMI'06). pp. 13--13. IEEE (2006)

\bibitem{li2022srfeat}
Li, L., Attaiki, S., Ovsjanikov, M.: Srfeat: Learning locally accurate and globally consistent non-rigid shape correspondence. In: International Conference on 3D Vision (3DV). IEEE (2022)

\bibitem{li2022learning}
Li, L., Donati, N., Ovsjanikov, M.: Learning multi-resolution functional maps with spectral attention for robust shape matching. NIPS  (2022)

\bibitem{li2017flame}
Li, T., Bolkart, T., Black, M.J., Li, H., Romero, J.: Learning a model of facial shape and expression from {4D} scans. ACM Transactions on Graphics (ToG)  \textbf{36}(6),  194:1--194:17 (2017)

\bibitem{litany2017deep}
Litany, O., Remez, T., Rodola, E., Bronstein, A., Bronstein, M.: Deep functional maps: Structured prediction for dense shape correspondence. In: ICCV (2017)

\bibitem{loper2015smpl}
Loper, M., Mahmood, N., Romero, J., Pons-Moll, G., Black, M.J.: Smpl: A skinned multi-person linear model. ACM Transactions on Graphics (ToG)  \textbf{34}(6),  1--16 (2015)

\bibitem{marin2020correspondence}
Marin, R., Rakotosaona, M.J., Melzi, S., Ovsjanikov, M.: Correspondence learning via linearly-invariant embedding. In: NeurIPS (2020)

\bibitem{melzi2019shrec}
Melzi, S., Marin, R., Rodol{\`a}, E., Castellani, U., Ren, J., Poulenard, A., Wonka, P., Ovsjanikov, M.: Shrec 2019: Matching humans with different connectivity. In: Eurographics Workshop on 3D Object Retrieval (2019)

\bibitem{melzi2019zoomout}
Melzi, S., Ren, J., Rodol{\`a}, E., Sharma, A., Wonka, P., Ovsjanikov, M.: Zoomout: spectral upsampling for efficient shape correspondence. ACM Transactions on Graphics (ToG)  \textbf{38}(6),  1--14 (2019)

\bibitem{ovsjanikov2012functional}
Ovsjanikov, M., Ben-Chen, M., Solomon, J., Butscher, A., Guibas, L.: Functional maps: a flexible representation of maps between shapes. ACM Transactions on Graphics (ToG)  \textbf{31}(4),  1--11 (2012)

\bibitem{ovsjanikov2010one}
Ovsjanikov, M., M{\'e}rigot, Q., M{\'e}moli, F., Guibas, L.: One point isometric matching with the heat kernel. In: Computer Graphics Forum. Wiley Online Library (2010)

\bibitem{pai2021fast}
Pai, G., Ren, J., Melzi, S., Wonka, P., Ovsjanikov, M.: Fast sinkhorn filters: Using matrix scaling for non-rigid shape correspondence with functional maps. In: CVPR (2021)

\bibitem{pierson2025diffumatch}
Pierson, E., Li, L., Dai, A., Ovsjanikov, M.: Diffumatch: Category-agnostic spectral diffusion priors for robust non-rigid shape matching. In: ICCV (2025)

\bibitem{pinkall1993computing}
Pinkall, U., Polthier, K.: Computing discrete minimal surfaces and their conjugates. Experimental mathematics  \textbf{2}(1),  15--36 (1993)

\bibitem{ren2021discrete}
Ren, J., Melzi, S., Wonka, P., Ovsjanikov, M.: Discrete optimization for shape matching. In: Computer Graphics Forum. Wiley Online Library (2021)

\bibitem{rodola2017partial}
Rodol{\`a}, E., Cosmo, L., Bronstein, M.M., Torsello, A., Cremers, D.: Partial functional correspondence. In: Computer Graphics Forum. Wiley Online Library (2017)

\bibitem{rodola2015point}
Rodol{\`a}, E., Moeller, M., Cremers, D.: Point-wise map recovery and refinement from functional correspondence. arXiv preprint arXiv:1506.05603  (2015)

\bibitem{rodola2017regularized}
Rodola, E., Moeller, M., Cremers, D.: Regularized pointwise map recovery from functional correspondence. In: Computer Graphics Forum. Wiley Online Library (2017)

\bibitem{roetzer2022scalable}
Roetzer, P., Swoboda, P., Cremers, D., Bernard, F.: A scalable combinatorial solver for elastic geometrically consistent 3d shape matching. In: CVPR (2022)

\bibitem{roufosse2019unsupervised}
Roufosse, J.M., Sharma, A., Ovsjanikov, M.: Unsupervised deep learning for structured shape matching. In: ICCV (2019)

\bibitem{sahilliouglu2020recent}
Sahillio{\u{g}}lu, Y.: Recent advances in shape correspondence. The Visual Computer  \textbf{36}(8),  1705--1721 (2020)

\bibitem{salti2014shot}
Salti, S., Tombari, F., Di~Stefano, L.: Shot: Unique signatures of histograms for surface and texture description. Computer Vision and Image Understanding  \textbf{125},  251--264 (2014)

\bibitem{sharma2020weakly}
Sharma, A., Ovsjanikov, M.: Weakly supervised deep functional maps for shape matching. NIPS  (2020)

\bibitem{sharp2020diffusionnet}
Sharp, N., Attaiki, S., Crane, K., Ovsjanikov, M.: Diffusionnet: Discretization agnostic learning on surfaces. arXiv preprint arXiv:2012.00888  (2020)

\bibitem{simeoni2025dinov3}
Sim{\'e}oni, O., Vo, H.V., Seitzer, M., Baldassarre, F., Oquab, M., Jose, C., Khalidov, V., Szafraniec, M., Yi, S., Ramamonjisoa, M., et~al.: Dinov3. arXiv preprint arXiv:2508.10104  (2025)

\bibitem{song20213d}
Song, C., Wei, J., Li, R., Liu, F., Lin, G.: 3d pose transfer with correspondence learning and mesh refinement. NIPS  (2021)

\bibitem{song2023unsupervised}
Song, C., Wei, J., Li, R., Liu, F., Lin, G.: Unsupervised 3d pose transfer with cross consistency and dual reconstruction. IEEE Transactions on Pattern Analysis and Machine Intelligence  (2023)

\bibitem{sun2023spatially}
Sun, M., Mao, S., Jiang, P., Ovsjanikov, M., Huang, R.: Spatially and spectrally consistent deep functional maps. In: ICCV (2023)

\bibitem{tam2012registration}
Tam, G.K., Cheng, Z.Q., Lai, Y.K., Langbein, F.C., Liu, Y., Marshall, D., Martin, R.R., Sun, X.F., Rosin, P.L.: Registration of 3d point clouds and meshes: A survey from rigid to nonrigid. IEEE transactions on visualization and computer graphics  \textbf{19}(7),  1199--1217 (2012)

\bibitem{trappolini2021shape}
Trappolini, G., Cosmo, L., Moschella, L., Marin, R., Melzi, S., Rodol{\`a}, E.: Shape registration in the time of transformers. NIPS  (2021)

\bibitem{van2011survey}
Van~Kaick, O., Zhang, H., Hamarneh, G., Cohen-Or, D.: A survey on shape correspondence. In: Computer Graphics Forum. Wiley Online Library (2011)

\bibitem{vestner2017product}
Vestner, M., Litman, R., Rodola, E., Bronstein, A., Cremers, D.: Product manifold filter: Non-rigid shape correspondence via kernel density estimation in the product space. In: CVPR (2017)

\bibitem{vigano2025nam}
Vigan{\`o}, G., Ovsjanikov, M., Melzi, S.: Nam: Neural adjoint maps for refining shape correspondences. ACM Transactions on Graphics (ToG)  \textbf{44}(4),  1--15 (2025)

\bibitem{wang2013image}
Wang, F., Huang, Q., Guibas, L.J.: Image co-segmentation via consistent functional maps. In: ICCV (2013)

\bibitem{wang2025kh}
Wang, W., Wei{\ss}berg, T., El~Amrani, N., Bernard, F.: kh: Symmetry understanding of 3d shapes via chirality disentanglement. In: ICCV (2025)

\bibitem{windheuser2011geometrically}
Windheuser, T., Schlickewei, U., Schmidt, F.R., Cremers, D.: Geometrically consistent elastic matching of 3d shapes: A linear programming solution. In: ICCV (2011)

\bibitem{xie2025echomatch}
Xie, Y., Ehm, V., Roetzer, P., El~Amrani, N., Gao, M., Bernard, F., Cremers, D.: Echomatch: Partial-to-partial shape matching via correspondence reflection. In: CVPR (2025)

\bibitem{zeng2021corrnet3d}
Zeng, Y., Qian, Y., Zhu, Z., Hou, J., Yuan, H., He, Y.: Corrnet3d: Unsupervised end-to-end learning of dense correspondence for 3d point clouds. In: CVPR (2021)

\bibitem{zhuravlev2025denoising}
Zhuravlev, A., L{\"a}hner, Z., Golyanik, V.: Denoising functional maps: Diffusion models for shape correspondence. In: CVPR (2025)

\end{thebibliography}

\clearpage
\setcounter{page}{1}
\title{Supplementary Material: } 

\titlerunning{Hyper-Network Neural Functional Maps for Unsupervised Robust 3D Shape Matching}

\author{Dongliang Cao\inst{1}\orcidlink{0000-0002-6505-6465} \and
Florian Bernard\inst{1}\orcidlink{0009-0008-1137-0003}}

\authorrunning{Dongliang Cao et al.}

\institute{University of Bonn}

\maketitle

In this supplementary document, we first provide a detailed description on deep functional map block~\cite{cao2023unsupervised} used in our framework (see~\cref{fig:pipeline_nfm}). Afterwards, we provide the implementation details of our method. Eventually, we show more qualitative results of our method.

\section{Deep Functional Map Block}
The overall pipeline of the deep functional map block is summarized in~\cref{sec:background}. In summary, the deep functional map block consists of three building blocks:
\begin{itemize}
    \item A Siamese feature extractor that processes both shapes to extract vertex-wise features, which are later used for both vertex-wise correspondence estimation and functional map computation.
    \item A vertex-wise correspondence prediction block that estimates point-wise correspondences based on feature similarity, see~\cref{eq:soft_corr}.
    \item A functional map solver that compute functional maps based on feature preservation assumption and additional structure properties, see~\cref{eq:fmap}.
\end{itemize}

During training, structure regularization is imposed on the functional maps computed from functional map solver to update the weights of the feature extractor. Specifically,  orthogonality, bijectivity regularization~\cite{roufosse2019unsupervised} is imposed on the computed functional maps $C_{\mathcal{MN}}$, i.e.,
\begin{equation}
    \label{eq:l_struct}
    L_{\mathrm{struct}} = \lambda_{\mathrm{bij}}L_{\mathrm{bij}} + \lambda_{\mathrm{orth}}L_{\mathrm{orth}}.
\end{equation}
The bijectivity regularization $L_{\mathrm{bij}}$ enforces the map from $\mathcal{M}$ through $\mathcal{N}$ back to $\mathcal{M}$ to be the identity map (and vice versa), i.e.,
    \begin{equation}
        \label{eq:bij}
        L_{\mathrm{bij}}=\left\|C_{\mathcal{MN}}C_{\mathcal{NM}}-I\right\|^{2}_{F}+\left\|C_{\mathcal{NM}}C_{\mathcal{MN}}-I\right\|^{2}_{F}.
    \end{equation}
The orthogonality regularization $L_{\mathrm{orth}}$ prompts locally area-preserving matchings for both matching directions, i.e.
    \begin{equation}
        \label{eq:orth}
        L_{\mathrm{orth}}=\left\|C_{\mathcal{MN}}^{\top}C_{\mathcal{MN}}-I\right\|^{2}_{F}+\left\|C_{\mathcal{NM}}^{\top}C_{\mathcal{NM}}-I\right\|^{2}_{F}.
    \end{equation} 

Together with the coupling regularization $L_{\mathrm{couple}}$ defined in~\cref{eq:couple} and our unsupervised spectral alignment loss defined in~\cref{eq:l_align}, the total loss can be defined as 
\begin{equation}
    \label{eq:l_total}
    L_{\mathrm{total}} = L_{\mathrm{struct}} + \lambda_{\mathrm{couple}}L_{\mathrm{couple}} + \lambda_{\mathrm{align}}L_{\mathrm{align}}.
\end{equation}

During inference, the vertex-wise correspondences can be obtained simply by nearest neighbor search in the feature space, see~\cref{eq:nn_search}. Moreover, additional test-time adaptation is used to further improve the final matching performance for each test shape pair individually. Specifically, it computes $L_{\mathrm{total}}$ defined in~\cref{eq:l_total} and update the feature extractor in an iterative way during inference. In this way, both functional map and vertex-wise correspondences are update simultaneously, in contrast to classical post-processing techniques~\cite{melzi2019zoomout,ren2021discrete}.

\section{Implementation Details}
Our deep shape matching block is built upon the architecture proposed in ULRSSM~\cite{cao2023unsupervised}. Specifically, we integrate their official implementation to serve as the backbone for our experiments for topological noisy and partial shape matching. For the multi-modal matching experiment in~\cref{subsec:multimodal}, we adapt this component to follow the SSMSM~\cite{cao2023self} framework. In this setting, we employ the self-supervised objectives introduced in~\cite{cao2023self} in conjunction with our proposed unsupervised spectral alignment loss. For the spectral representation, we utilize the first 128 eigenfunctions of the Laplace-Beltrami Operator (LBO) as our basis (i.e., $k=128$). Our ViT-based hyper-network is adapted from the official PyTorch implementation with a patch size of 16. To maintain consistency with our spectral resolution, we set the latent feature dimension to 128 for both the multi-head self-attention (MSA) and the MLP blocks. The transformer encoder consists of 8 layers, each utilizing 8 attention heads to capture global spectral dependencies. We set the loss weight $\lambda_{\mathrm{align}}$ of our proposed spectral alignment loss to be $1e^{-4}$. We use the same setting across all different experiments. 

\section{More Ablation Studies}
\subsection{Sign Ambiguity}
We stress-test by randomly flipping $p = 50\%$ of eigenvector signs at test time on HOLES dataset: geodesic error remains \emph{almost unchanged} (before 4.74 vs after 4.82). This is because NFM only learns a small non-linear residual on top of the linear functional map $C$ (\cref{eq:nfm}), which itself absorbs the sign-induced transformations/ambiguities.

\subsection{Spectral dimension $k$}
We evaluate the impact of spectral dimension $k$ on SHREC16 HOLES dataset. In this experiment, $k\!\in\!\{32,64,96,160\}$ gives 9.5/6.9/5.7/5.0 geodesic error ($\times 100$) and thus the default $k\!=\!128$ is optimal with 4.7 geodesic error ($\times 100$).

\section{More Qualitative Results}
In the next figures, we provide additional qualitative results of our method corresponding to the quantitative results reported in the main text. 

\begin{figure}[bht!]
    \centering
\def\rowOnecolumnOne{kid00-kid04}
\def\rowOnecolumnTwo{kid00-kid25}
\def\rowOnecolumnThree{kid00-kid18}
\def\rowOnecolumnFour{kid00-kid07}
\def\rowOnecolumnFive{kid00-kid24}
\def\rowOnecolumnSix{kid00-kid04}

\def\pathShrecNT{figs/ours/topkids/}
\def\hspaceCols{0.13cm}
\def\wspaceRows{0cm}
\def\height{3.1cm}
\def\width{3.0cm}
\def\heightT{\height}
\def\widthT{\width}
\begin{tabular}{cccccc}%
    \setlength{\tabcolsep}{0pt} 
    {\scriptsize Source} & & & & & \\
    \vspace{\wspaceRows}
    \hspace{\hspaceCols}
    \includegraphics[height=\heightT, width=\widthT]{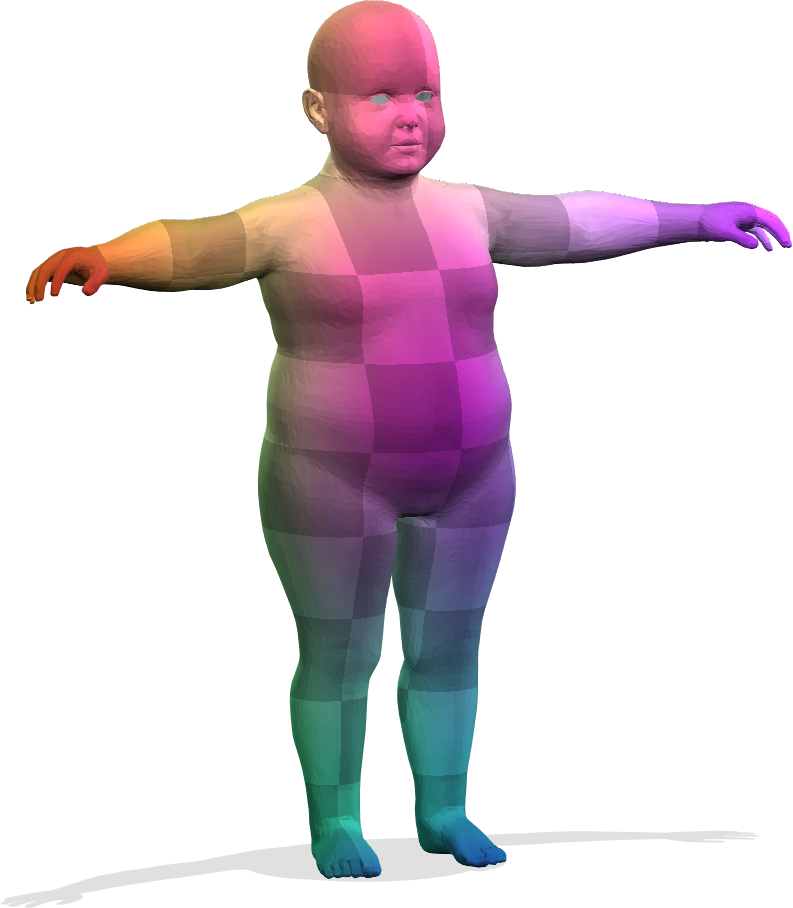}&
    \hspace{\hspaceCols}
    \includegraphics[height=\heightT, width=\widthT]{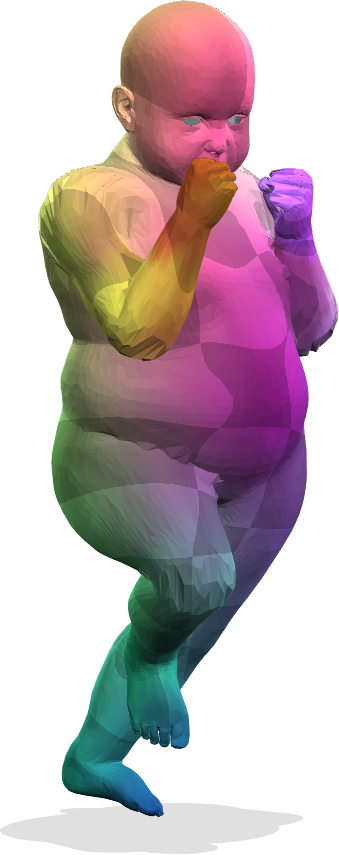}&
    \hspace{\hspaceCols}
    \includegraphics[height=\heightT, width=\widthT]{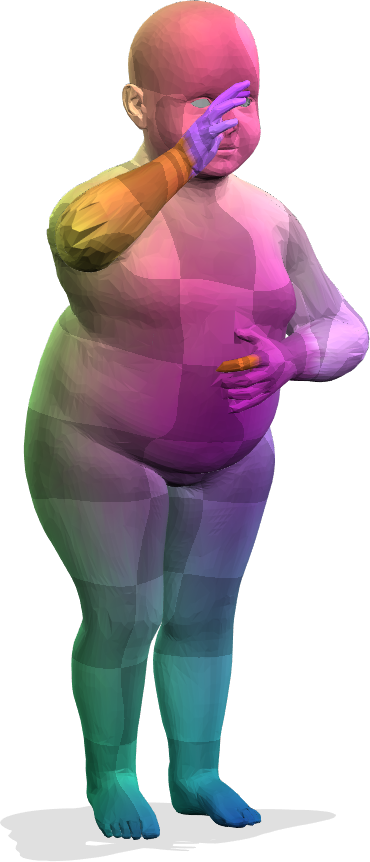}&
    \hspace{\hspaceCols}
    \includegraphics[height=\heightT, width=\widthT]{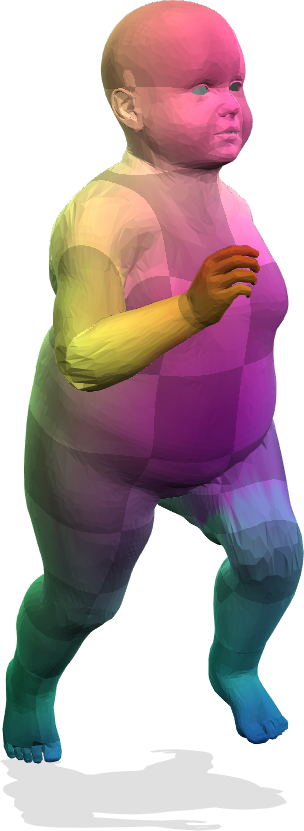}&
    \hspace{\hspaceCols}
    \includegraphics[height=2.8cm, width=\widthT]{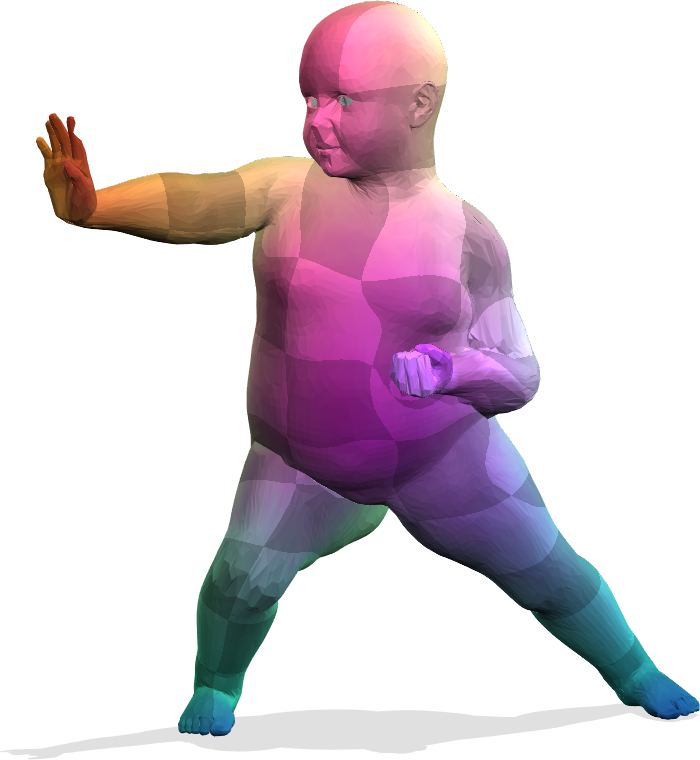}
    &
    \hspace{\hspaceCols}
    \includegraphics[height=\heightT, width=\widthT]{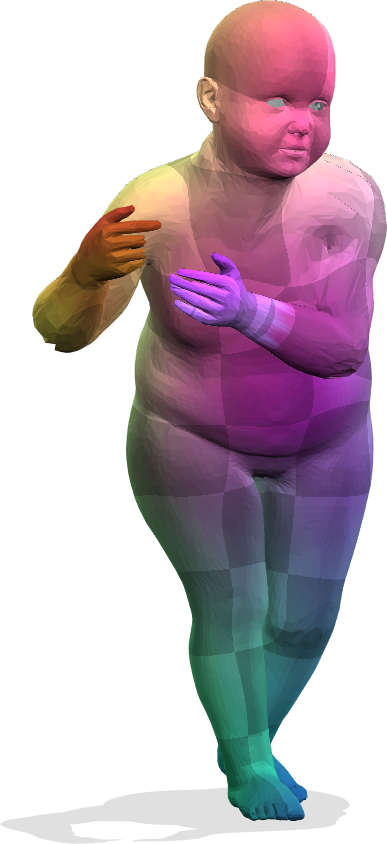}
    \\
\end{tabular}
    \caption{\textbf{Qualitative results of our method on TOPKIDS.} Our method obtains accurate correspondences for shapes with topological noise.}
    \label{fig:topkids_sup}
\end{figure}

\begin{figure}[bht!]
    \centering
\def\rowOnecolumnOne{10-8}
\def\rowOnecolumnTwo{10-8}
\def\rowOnecolumnThree{10-12}
\def\rowOnecolumnFour{10-16}
\def\rowOnecolumnFive{10-22}

\def\rowTwocolumnTwo{10-32}
\def\rowTwocolumnThree{10-36}
\def\rowTwocolumnFour{10-43}
\def\rowTwocolumnFive{10-44}

\def\pathShrecNT{figs/ours/shrec19/}
\def\hspaceCols{0.13cm}
\def\wspaceRows{0cm}
\def\height{3.1cm}
\def\width{3.0cm}
\def\heightT{\height}
\def\widthT{\width}
\begin{tabular}{ccccc}%
    \setlength{\tabcolsep}{0pt} 
    {\scriptsize Source} & & & &  \\
    \vspace{\wspaceRows}
    \hspace{\hspaceCols}
    \includegraphics[height=\heightT, width=\widthT]{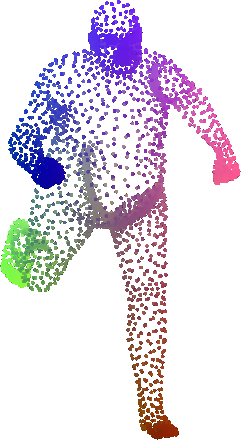}&
    \hspace{\hspaceCols}
    \includegraphics[height=\heightT, width=\widthT]{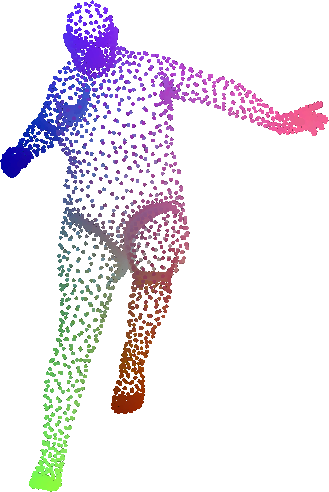}&
    \hspace{\hspaceCols}
    \includegraphics[height=\heightT, width=\widthT]{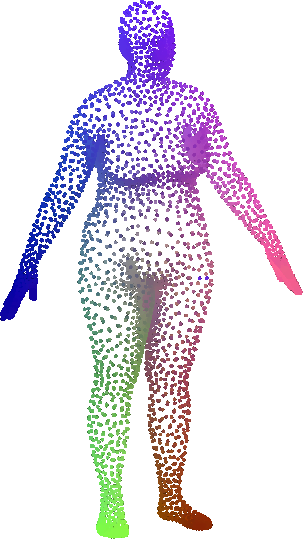}&
    \hspace{\hspaceCols}
    \includegraphics[height=\heightT, width=\widthT]{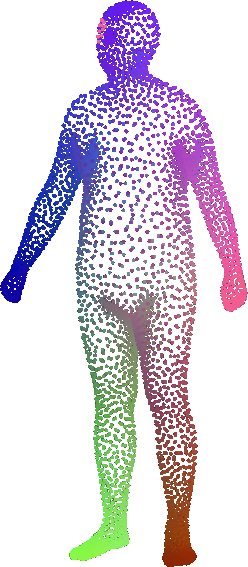}&
    \hspace{\hspaceCols}
    \includegraphics[height=\heightT, width=\widthT]{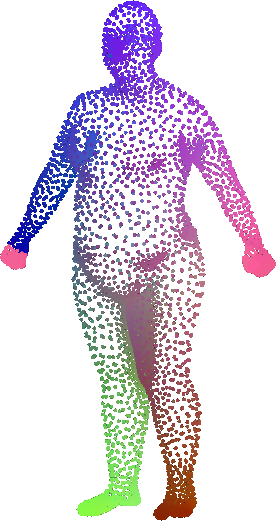}
    \\
    \hspace{\hspaceCols}
    &
    \hspace{\hspaceCols}
    \includegraphics[height=2.8cm, width=\widthT]{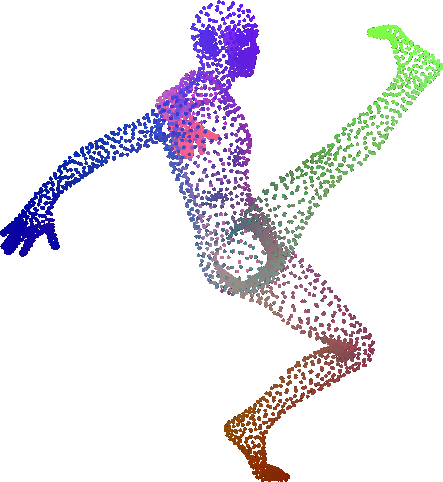}&
    \hspace{\hspaceCols}
    \includegraphics[height=\heightT, width=\widthT]{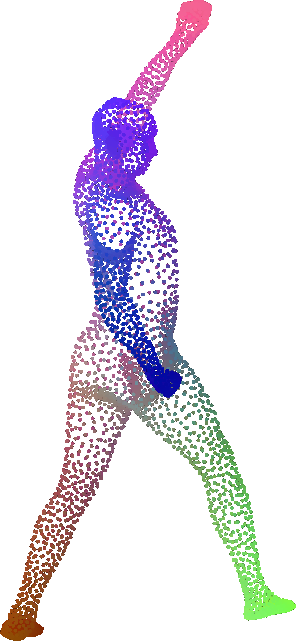}&
    \hspace{\hspaceCols}
    \includegraphics[height=\heightT, width=\widthT]{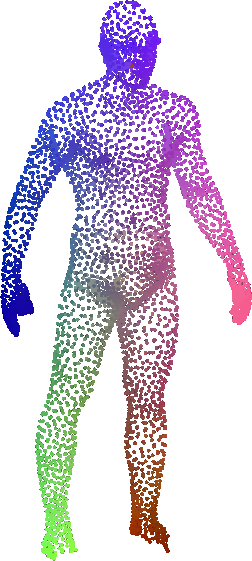}&
    \hspace{\hspaceCols}
    \includegraphics[height=2.8cm, width=\widthT]{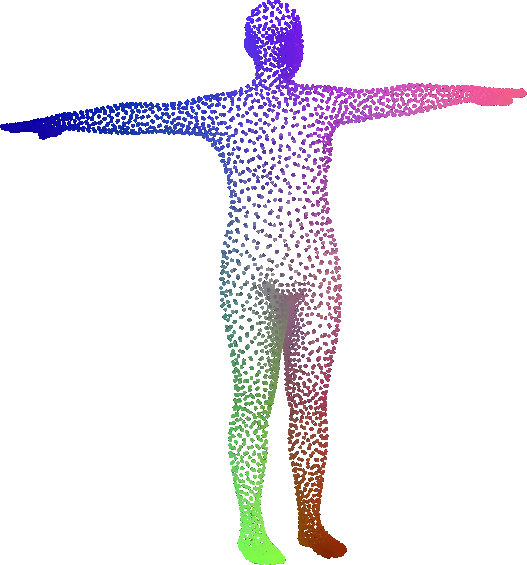}
    \\
\end{tabular}
    \caption{\textbf{Qualitative results of our method on SHREC19.} Our method combined with SSMSM obtains accurate correspondences for shapes represented as point clouds.}
    \label{fig:shrec19_sup}
\end{figure}

\begin{figure}[bht!]
    \centering
\def\rowOnecolumnOne{cat-cuts_cat_shape_11}
\def\rowOnecolumnTwo{cat-cuts_cat_shape_11}
\def\rowOnecolumnThree{cat-cuts_cat_shape_13}
\def\rowOnecolumnFour{cat-cuts_cat_shape_4}
\def\rowOnecolumnFive{cat-cuts_cat_shape_9}
\def\rowTwocolumnTwo{cat-cuts_cat_shape_6}
\def\rowTwocolumnThree{cat-cuts_cat_shape_22}
\def\rowTwocolumnFour{cat-cuts_cat_shape_26}
\def\rowTwocolumnFive{cat-cuts_cat_shape_28}

\def\rowThreecolumnOne{michael-cuts_michael_shape_11}
\def\rowThreecolumnTwo{michael-cuts_michael_shape_19}
\def\rowThreecolumnThree{michael-cuts_michael_shape_39}
\def\rowThreecolumnFour{michael-cuts_michael_shape_14}
\def\rowThreecolumnFive{michael-cuts_michael_shape_15}
\def\rowFourcolumnTwo{michael-cuts_michael_shape_38}
\def\rowFourcolumnThree{michael-cuts_michael_shape_13}
\def\rowFourcolumnFour{michael-cuts_michael_shape_23}
\def\rowFourcolumnFive{michael-cuts_michael_shape_26}

\def\rowFivecolumnOne{dog-cuts_dog_shape_1}
\def\rowFivecolumnTwo{dog-cuts_dog_shape_19}
\def\rowFivecolumnThree{dog-cuts_dog_shape_2}
\def\rowFivecolumnFour{dog-cuts_dog_shape_6}
\def\rowFivecolumnFive{dog-cuts_dog_shape_8}
\def\rowSixcolumnTwo{dog-cuts_dog_shape_9}
\def\rowSixcolumnThree{dog-cuts_dog_shape_10}
\def\rowSixcolumnFour{dog-cuts_dog_shape_12}
\def\rowSixcolumnFive{dog-cuts_dog_shape_15}

\def\pathShrecNT{figs/ours/cuts/}
\def\hspaceCols{0.13cm}
\def\wspaceRows{0cm}
\def\height{2.5cm}
\def\width{2.0cm}
\def\heightT{\height}
\def\widthT{\width}
\begin{tabular}{ccccc}%
    \setlength{\tabcolsep}{0pt} 
    {\scriptsize Source} & & & &  \\
    \vspace{\wspaceRows}
    \hspace{\hspaceCols}
    \includegraphics[height=\heightT, width=\widthT]{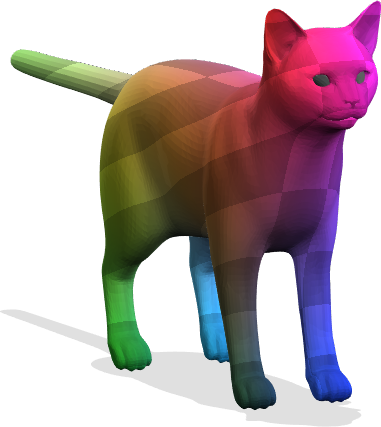}&
    \hspace{\hspaceCols}
    \includegraphics[height=\heightT, width=\widthT]{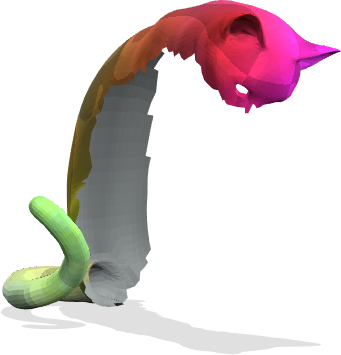}&
    \hspace{\hspaceCols}
    \includegraphics[height=\heightT, width=\widthT]{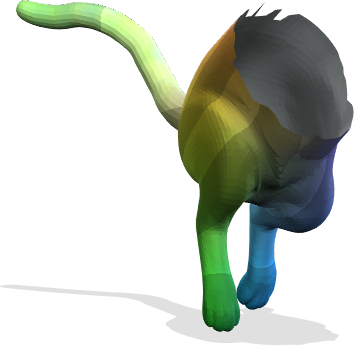}&
    \hspace{\hspaceCols}
    \includegraphics[height=\heightT, width=1.6cm]{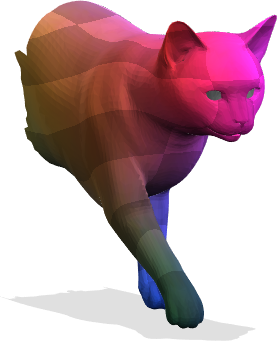}&
    \hspace{\hspaceCols}
    \includegraphics[height=\heightT, width=1.6cm]{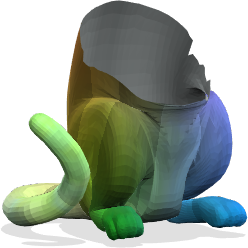}
    \\
    &
    \hspace{\hspaceCols}
    \includegraphics[height=\heightT, width=\widthT]{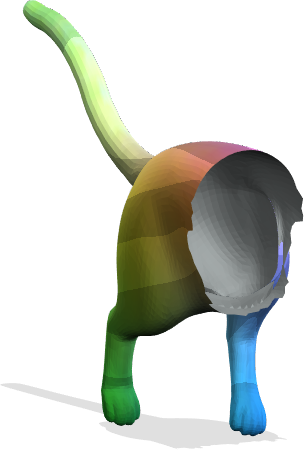}&
    \hspace{\hspaceCols}
    \includegraphics[height=\heightT, width=\widthT]{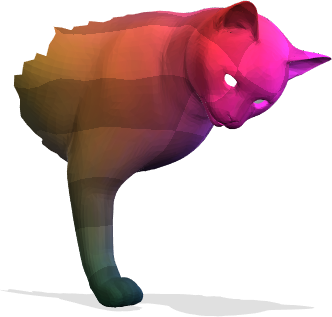}&
    \hspace{\hspaceCols}
    \includegraphics[height=\heightT, width=\widthT]{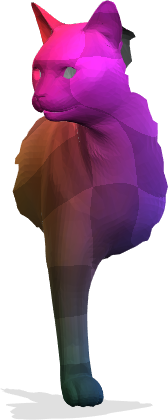}&
    \hspace{\hspaceCols}
    \includegraphics[height=\heightT, width=1.6cm]{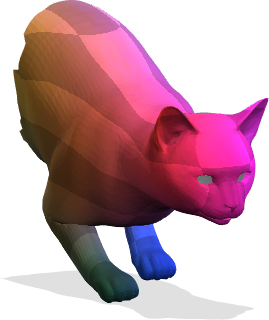}
    \\
    {\scriptsize Source} & & & &  \\
    \vspace{\wspaceRows}
    \hspace{\hspaceCols}
    \includegraphics[height=2.6cm, width=3.0cm]{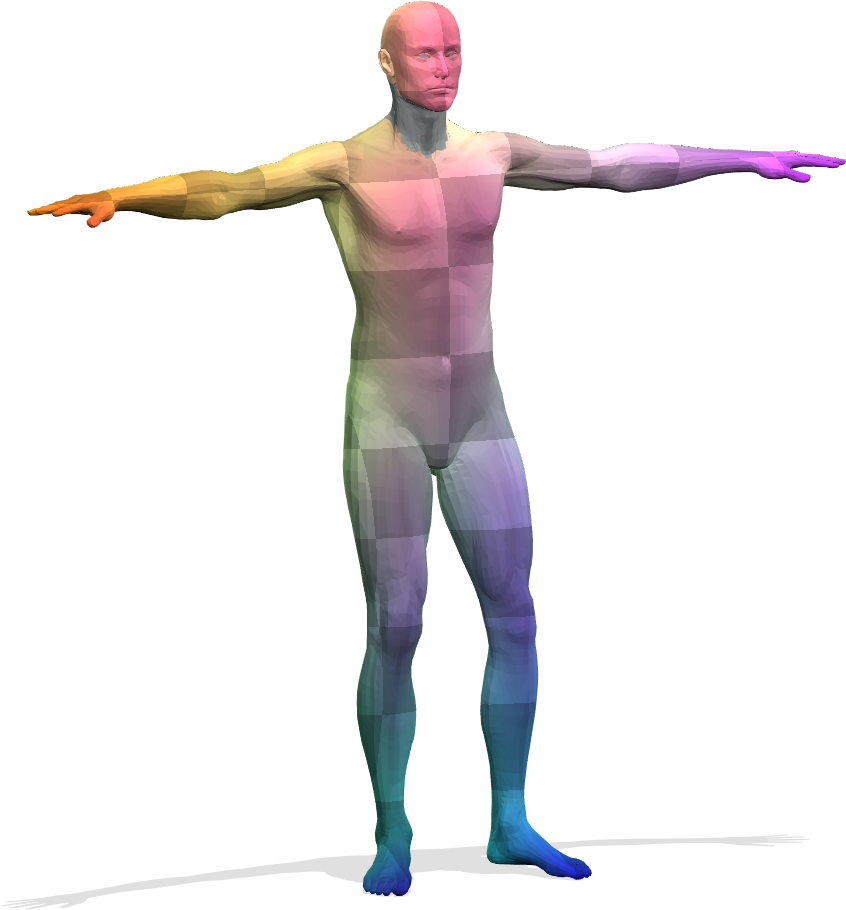}&
    \hspace{\hspaceCols}
    \includegraphics[height=2.6cm, width=2.3cm]{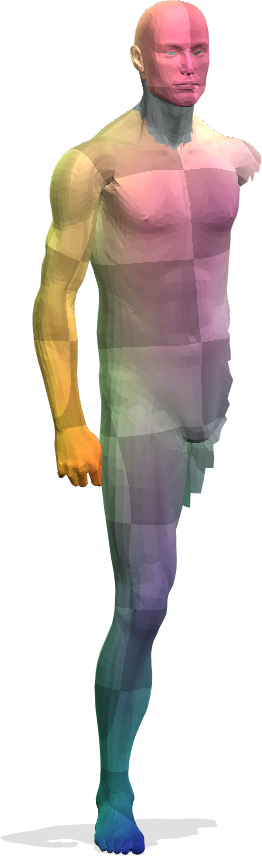}&
    \hspace{\hspaceCols}
    \includegraphics[height=2.6cm, width=2.3cm]{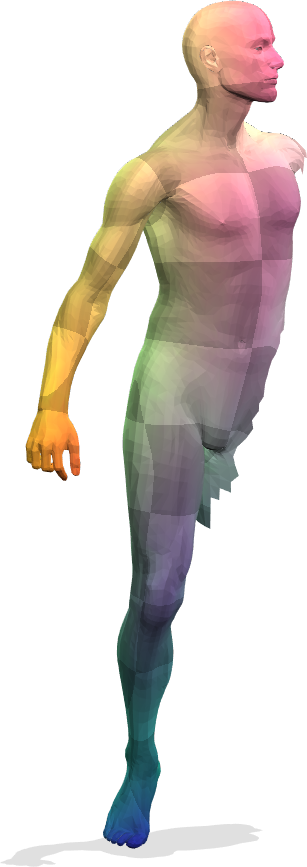}&
    \hspace{\hspaceCols}
    \includegraphics[height=2.6cm, width=2.3cm]{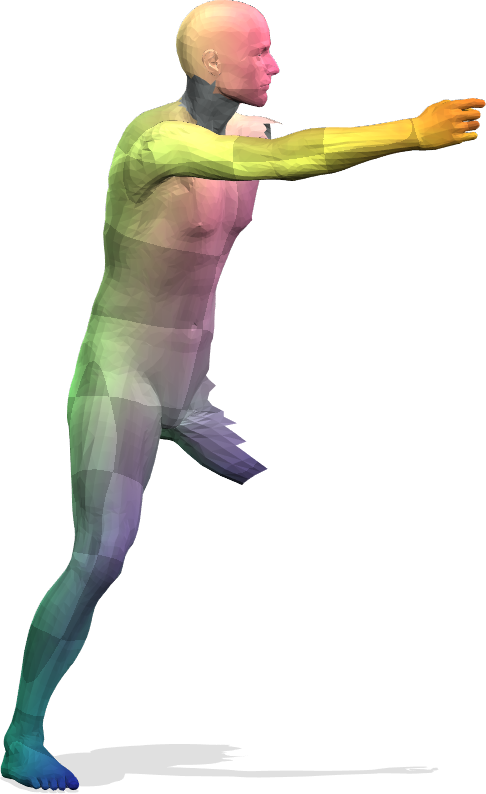}&
    \hspace{\hspaceCols}
    \includegraphics[height=\heightT, width=\widthT]{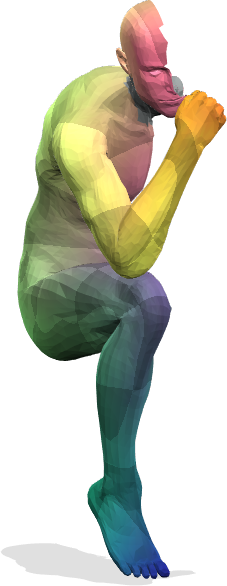}
    \\
    &
    \hspace{\hspaceCols}
    \includegraphics[height=\heightT, width=1.6cm]{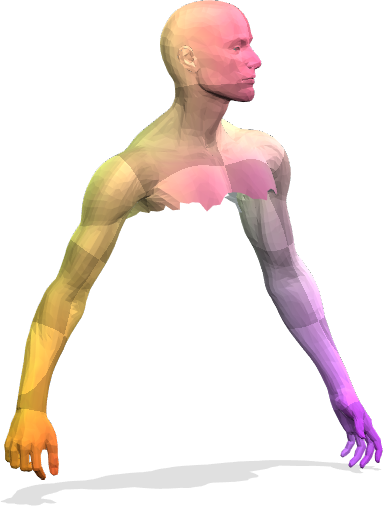}&
    \hspace{\hspaceCols}
    \includegraphics[height=\heightT, width=1.6cm]{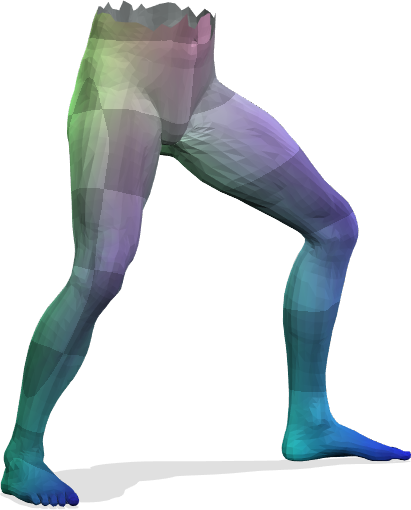}&
    \hspace{\hspaceCols}
    \includegraphics[height=\heightT, width=1.6cm]{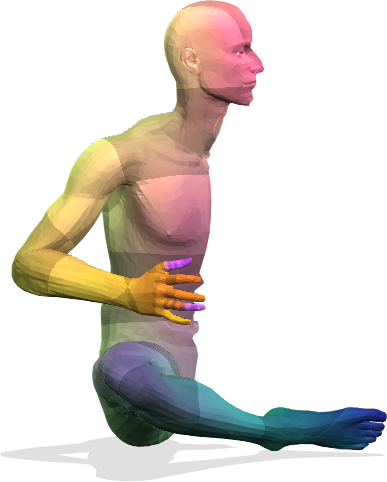}&
    \hspace{\hspaceCols}
    \includegraphics[height=2.0cm, width=1.6cm]{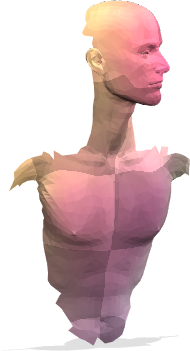}
    \\
    {\scriptsize Source} & & & &  \\
    \vspace{\wspaceRows}
    \hspace{\hspaceCols}
    \includegraphics[height=\heightT, width=1.5cm]{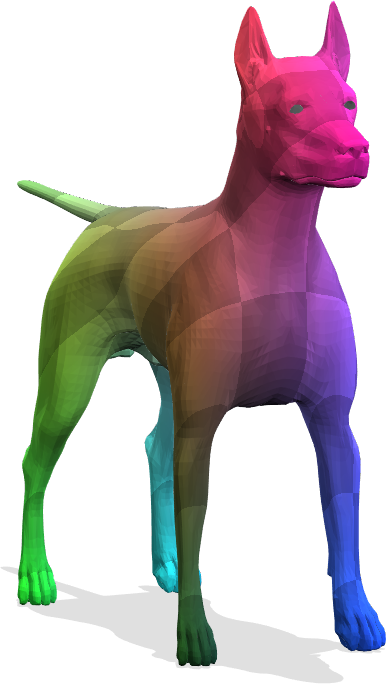}&
    \hspace{\hspaceCols}
    \includegraphics[height=\heightT, width=\widthT]{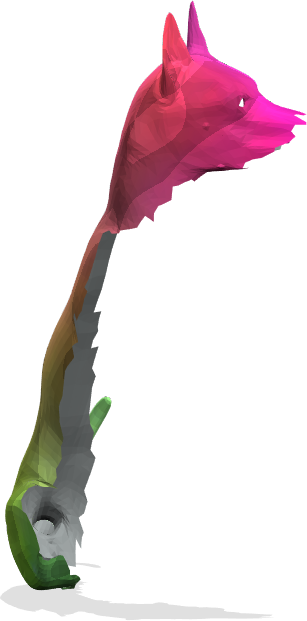}&
    \hspace{\hspaceCols}
    \includegraphics[height=\heightT, width=\widthT]{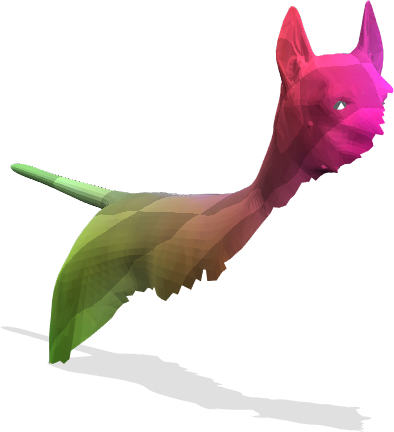}&
    \hspace{\hspaceCols}
    \includegraphics[height=\heightT, width=2.3cm]{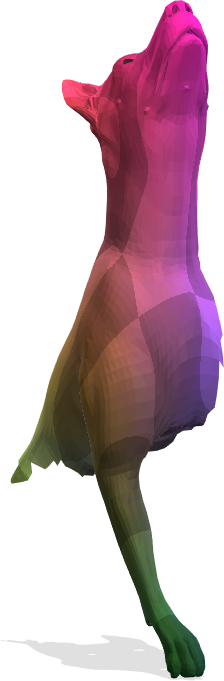}&
    \hspace{\hspaceCols}
    \includegraphics[height=\heightT, width=1.6cm]{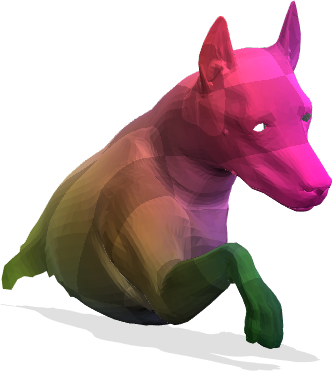}
    \\
    &
    \hspace{\hspaceCols}
    \includegraphics[height=\heightT, width=\widthT]{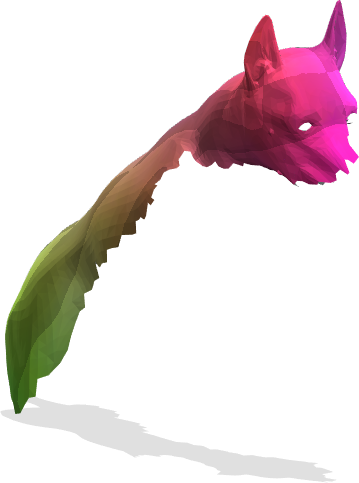}&
    \hspace{\hspaceCols}
    \includegraphics[height=\heightT, width=\widthT]{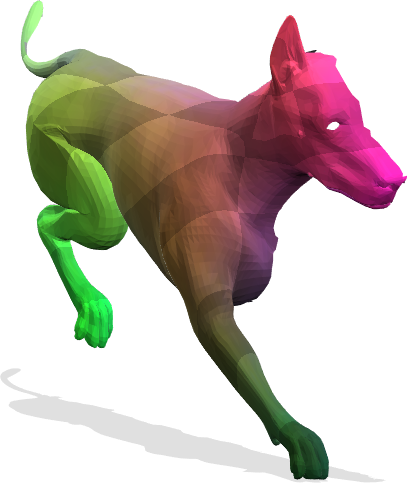}&
    \hspace{\hspaceCols}
    \includegraphics[height=\heightT, width=\widthT]{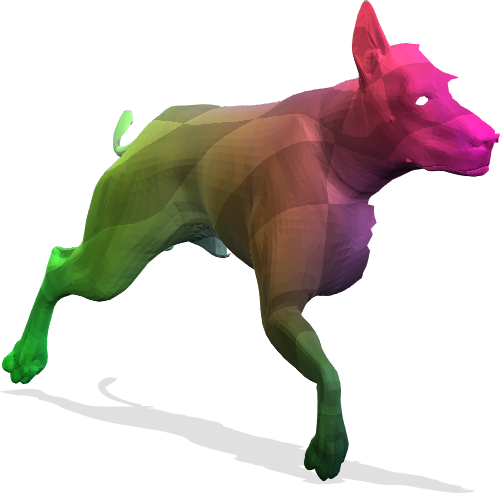}&
    \hspace{\hspaceCols}
    \includegraphics[height=\heightT, width=\widthT]{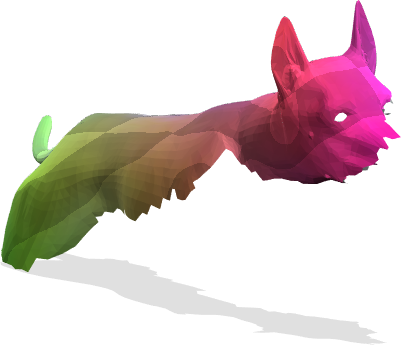}
    \\
\end{tabular}
    \caption{\textbf{Qualitative results of our method on SHREC16 CUTS.} Our method obtains accurate correspondences for partial shapes.}
    \label{fig:cuts_sup}
\end{figure}

\begin{figure}[bht!]
    \centering
\def\rowOnecolumnOne{cat-holes_cat_shape_3}
\def\rowOnecolumnTwo{cat-holes_cat_shape_3}
\def\rowOnecolumnThree{cat-holes_cat_shape_7}
\def\rowOnecolumnFour{cat-holes_cat_shape_9}
\def\rowOnecolumnFive{cat-holes_cat_shape_10}
\def\rowTwocolumnTwo{cat-holes_cat_shape_11}
\def\rowTwocolumnThree{cat-holes_cat_shape_15}
\def\rowTwocolumnFour{cat-holes_cat_shape_16}
\def\rowTwocolumnFive{cat-holes_cat_shape_18}

\def\rowThreecolumnOne{michael-holes_michael_shape_7}
\def\rowThreecolumnTwo{michael-holes_michael_shape_51}
\def\rowThreecolumnThree{michael-holes_michael_shape_10}
\def\rowThreecolumnFour{michael-holes_michael_shape_13}
\def\rowThreecolumnFive{michael-holes_michael_shape_18}
\def\rowFourcolumnTwo{michael-holes_michael_shape_22}
\def\rowFourcolumnThree{michael-holes_michael_shape_26}
\def\rowFourcolumnFour{michael-holes_michael_shape_28}
\def\rowFourcolumnFive{michael-holes_michael_shape_32}

\def\rowFivecolumnOne{horse-holes_horse_shape_2}
\def\rowFivecolumnTwo{horse-holes_horse_shape_2}
\def\rowFivecolumnThree{horse-holes_horse_shape_3}
\def\rowFivecolumnFour{horse-holes_horse_shape_4}
\def\rowFivecolumnFive{horse-holes_horse_shape_5}
\def\rowSixcolumnTwo{horse-holes_horse_shape_6}
\def\rowSixcolumnThree{horse-holes_horse_shape_9}
\def\rowSixcolumnFour{horse-holes_horse_shape_10}
\def\rowSixcolumnFive{horse-holes_horse_shape_12}

\def\pathShrecNT{figs/ours/holes/}
\def\hspaceCols{0.13cm}
\def\wspaceRows{0cm}
\def\height{2.5cm}
\def\width{2.0cm}
\def\heightT{\height}
\def\widthT{\width}
\begin{tabular}{ccccc}%
    \setlength{\tabcolsep}{0pt} 
    {\scriptsize Source} & & & &  \\
    \vspace{\wspaceRows}
    \hspace{\hspaceCols}
    \includegraphics[height=\heightT, width=\widthT]{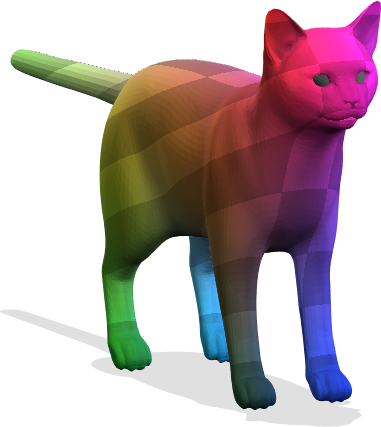}&
    \hspace{\hspaceCols}
    \includegraphics[height=\heightT, width=\widthT]{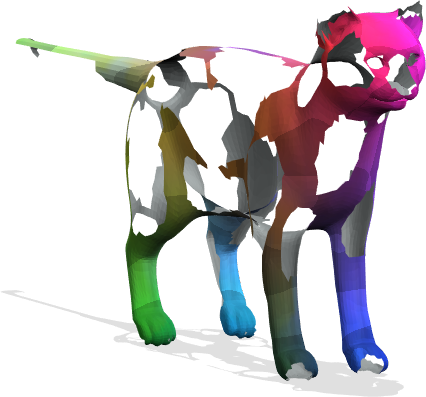}&
    \hspace{\hspaceCols}
    \includegraphics[height=\heightT, width=\widthT]{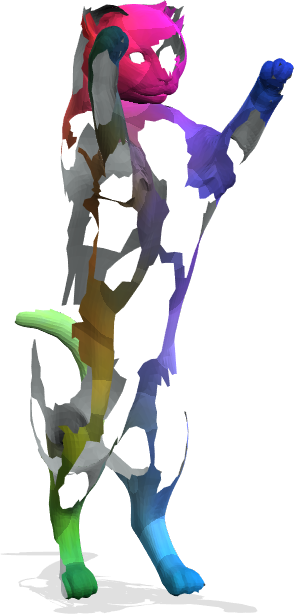}&
    \hspace{\hspaceCols}
    \includegraphics[height=\heightT, width=1.6cm]{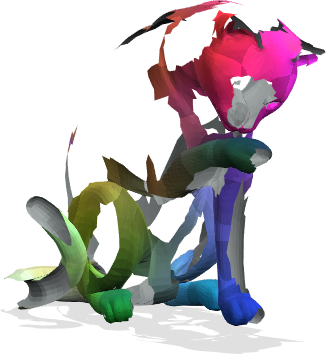}&
    \hspace{\hspaceCols}
    \includegraphics[height=\heightT, width=1.6cm]{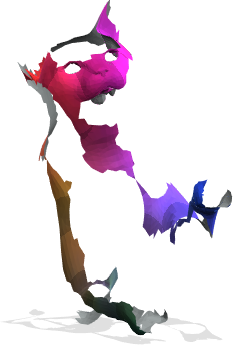}
    \\
    &
    \hspace{\hspaceCols}
    \includegraphics[height=\heightT, width=\widthT]{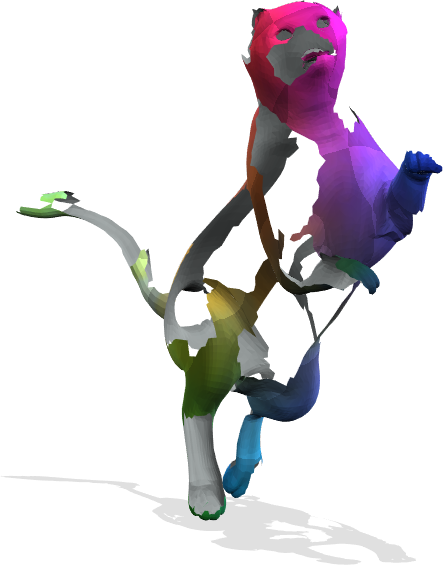}&
    \hspace{\hspaceCols}
    \includegraphics[height=\heightT, width=1.2cm]{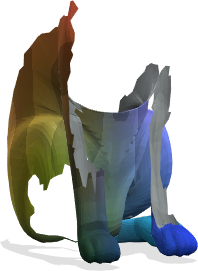}&
    \hspace{\hspaceCols}
    \includegraphics[height=\heightT, width=1.6cm]{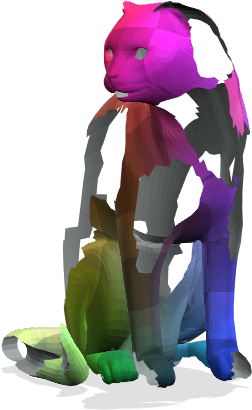}&
    \hspace{\hspaceCols}
    \includegraphics[height=\heightT, width=1.6cm]{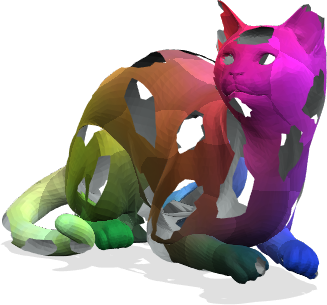}
    \\
    {\scriptsize Source} & & & &  \\
    \vspace{\wspaceRows}
    \hspace{\hspaceCols}
    \includegraphics[height=2.6cm, width=3.0cm]{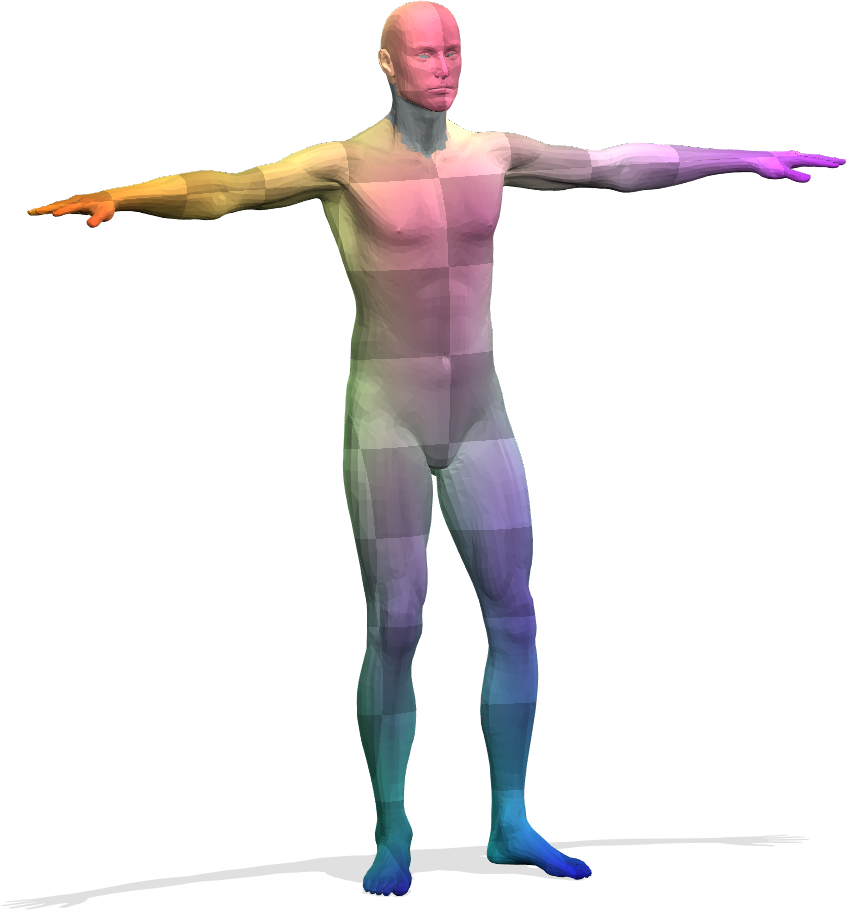}&
    \hspace{\hspaceCols}
    \includegraphics[height=2.6cm, width=2.3cm]{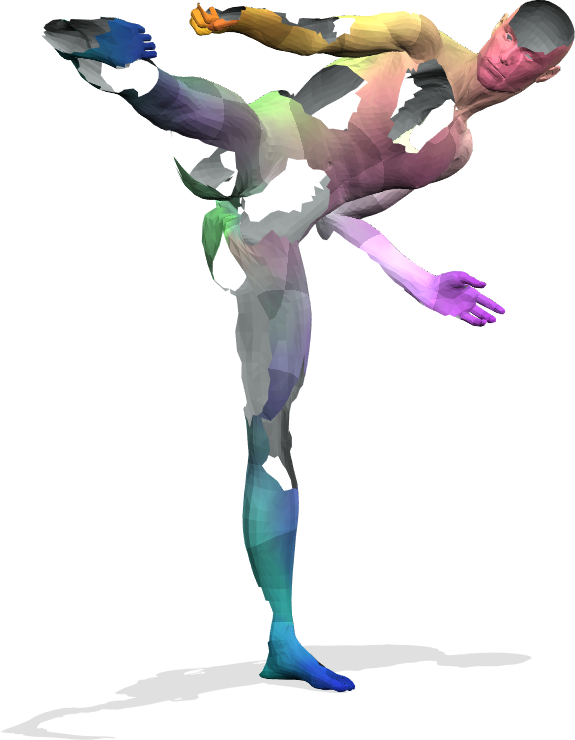}&
    \hspace{\hspaceCols}
    \includegraphics[height=2.6cm, width=2.3cm]{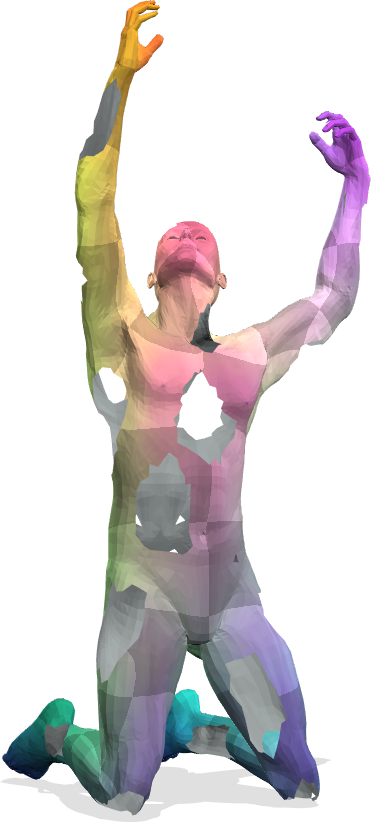}&
    \hspace{\hspaceCols}
    \includegraphics[height=2.6cm, width=2.3cm]{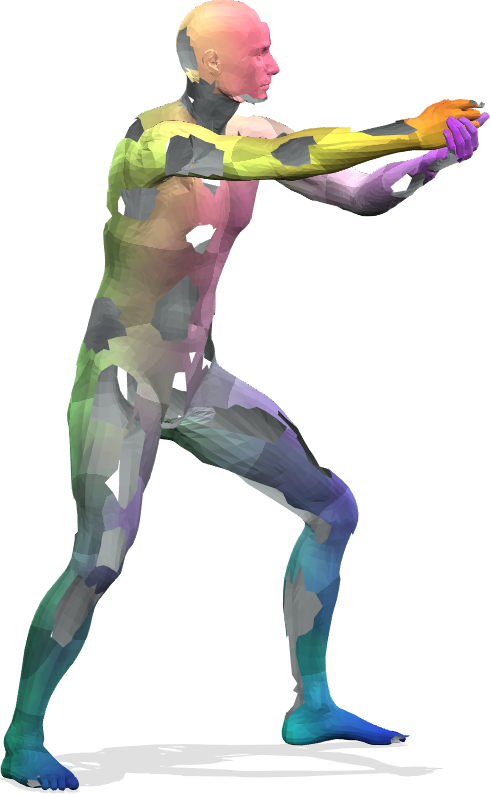}&
    \hspace{\hspaceCols}
    \includegraphics[height=2.6cm, width=2.3cm]{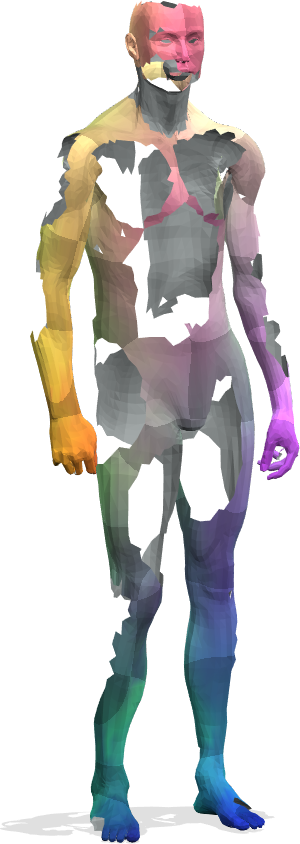}
    \\
    &
    \hspace{\hspaceCols}
    \includegraphics[height=\heightT, width=1.6cm]{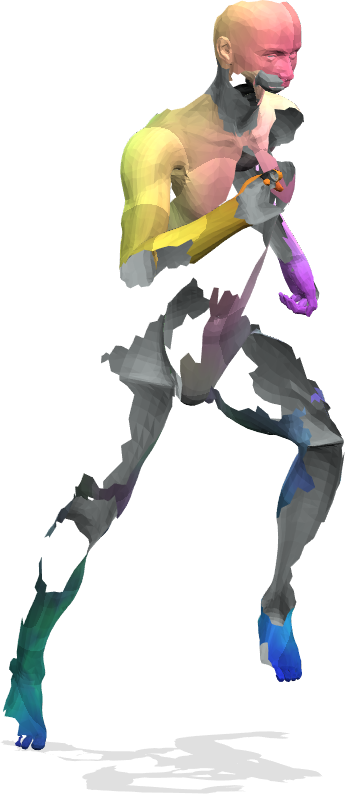}&
    \hspace{\hspaceCols}
    \includegraphics[height=\heightT, width=1.6cm]{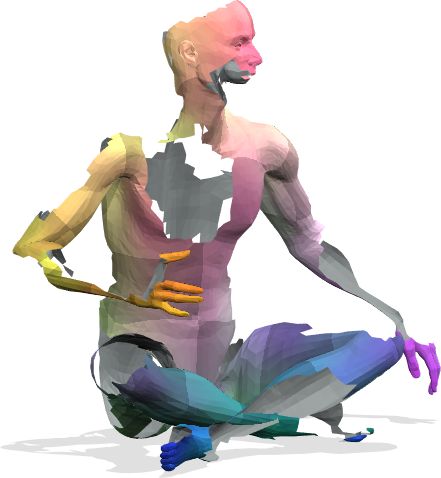}&
    \hspace{\hspaceCols}
    \includegraphics[height=\heightT, width=1.6cm]{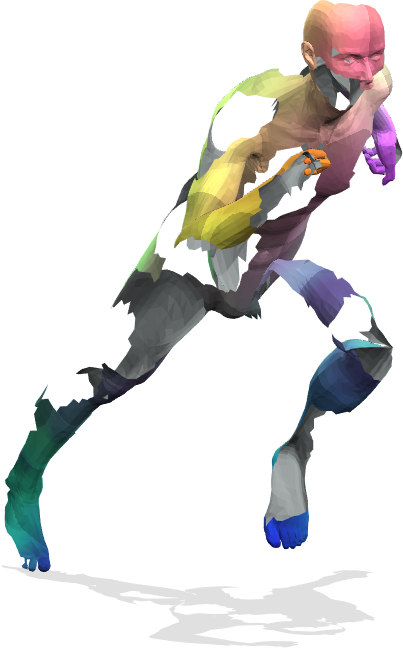}&
    \hspace{\hspaceCols}
    \includegraphics[height=2.6cm, width=2.3cm]{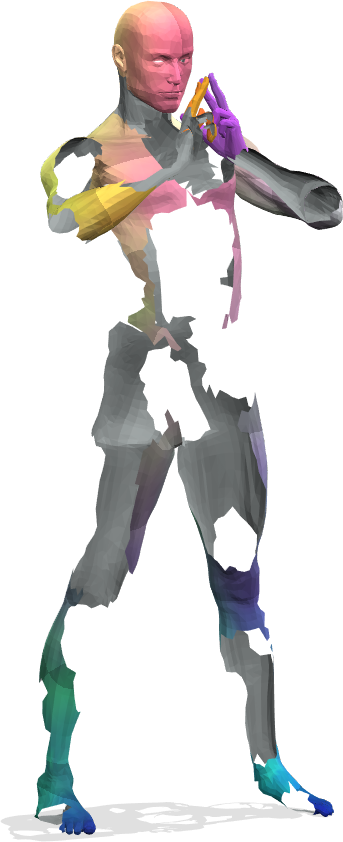}
    \\
    {\scriptsize Source} & & & &  \\
    \vspace{\wspaceRows}
    \hspace{\hspaceCols}
    \includegraphics[height=\heightT, width=1.5cm]{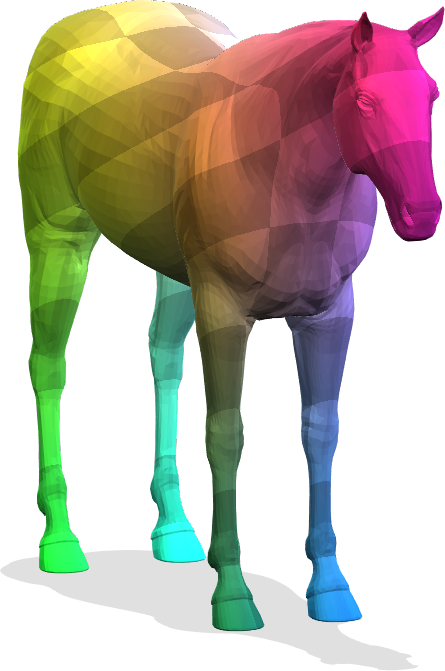}&
    \hspace{\hspaceCols}
    \includegraphics[height=\heightT, width=\widthT]{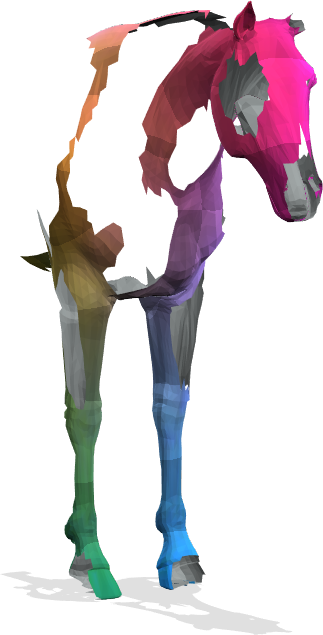}&
    \hspace{\hspaceCols}
    \includegraphics[height=\heightT, width=\widthT]{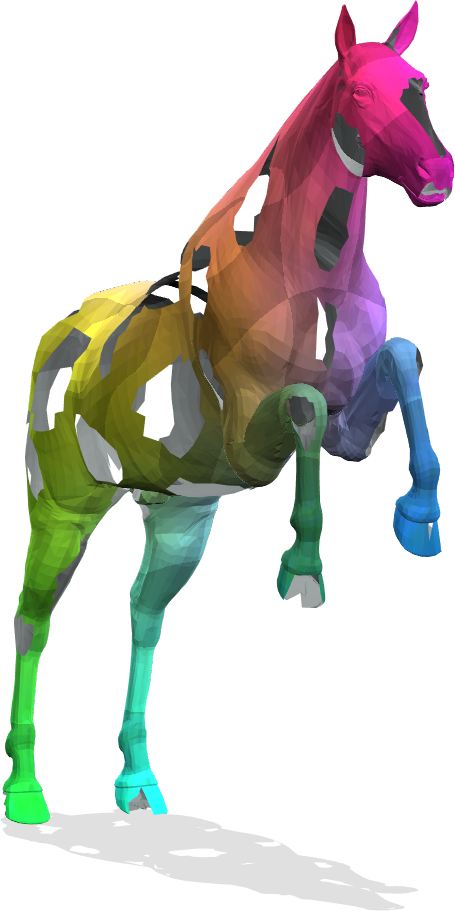}&
    \hspace{\hspaceCols}
    \includegraphics[height=\heightT, width=2.3cm]{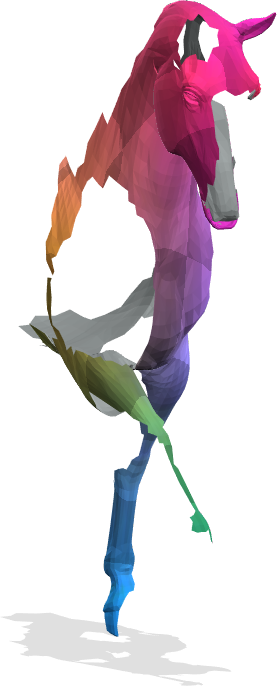}&
    \hspace{\hspaceCols}
    \includegraphics[height=\heightT, width=1.6cm]{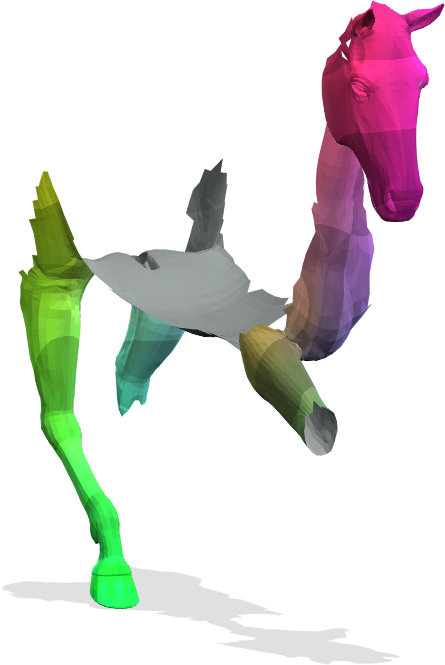}
    \\
    &
    \hspace{\hspaceCols}
    \includegraphics[height=\heightT, width=\widthT]{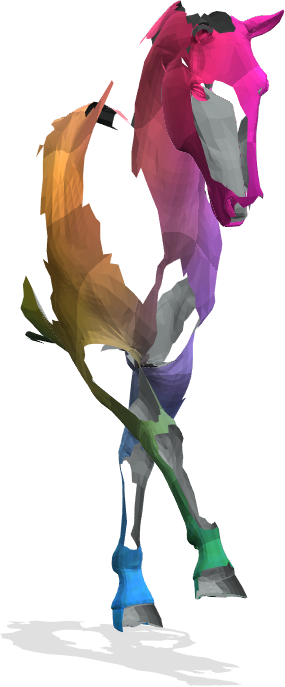}&
    \hspace{\hspaceCols}
    \includegraphics[height=\heightT, width=\widthT]{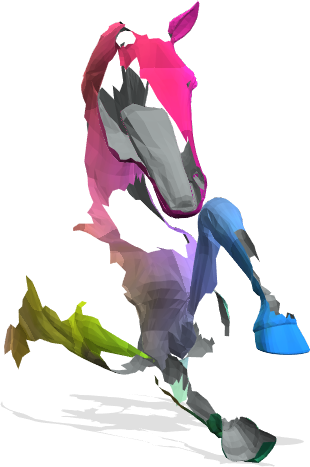}&
    \hspace{\hspaceCols}
    \includegraphics[height=\heightT, width=\widthT]{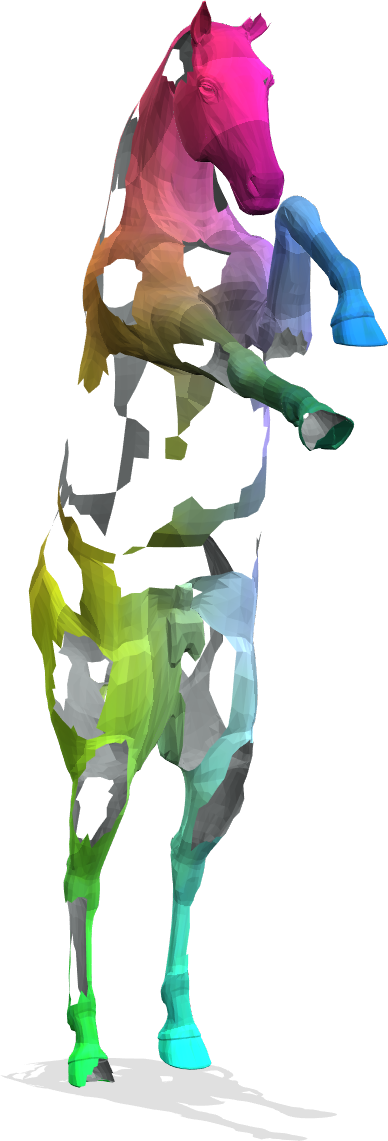}&
    \hspace{\hspaceCols}
    \includegraphics[height=\heightT, width=\widthT]{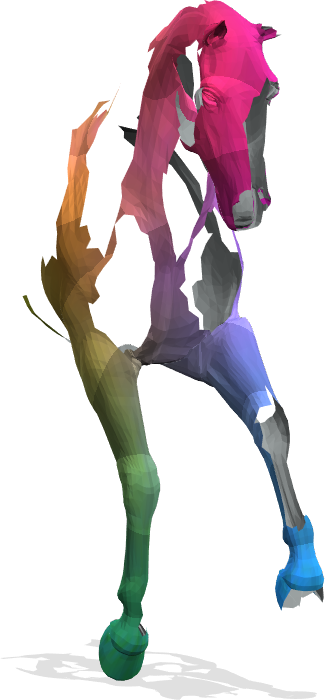}
    \\
\end{tabular}
    \caption{\textbf{Qualitative results of our method on SHREC16 HOLES.} Our method obtains accurate correspondences for shapes with extreme partiality.}
    \label{fig:holes_sup}
\end{figure}

\end{document}